%% file: neurips_2026.tex
\documentclass{article}
\usepackage{graphicx}
\usepackage{microtype}
\usepackage{graphicx}
\usepackage{subcaption}
\usepackage{booktabs}
\usepackage{booktabs}
\usepackage{multirow}
\usepackage{graphicx}
\usepackage{booktabs}
\usepackage{tabularx}
\usepackage{array}
\usepackage{makecell}
\usepackage{siunitx}
\usepackage{tcolorbox}
\usepackage{adjustbox}
\usepackage{enumitem}
\usepackage{hyperref}

\usepackage{amsmath}
\usepackage{amssymb}
\usepackage{mathtools}
\usepackage{amsthm}
\usepackage{hyperref}

\usepackage{etoolbox}

\usepackage{subcaption}
\usepackage[capitalize,noabbrev]{cleveref}
\theoremstyle{plain}

\theoremstyle{definition}

\theoremstyle{remark}

\PassOptionsToPackage{numbers,compress}{natbib}
\usepackage[preprint]{neurips_2026}

\usepackage{hyperref}

\usepackage[utf8]{inputenc} 
\usepackage[T1]{fontenc}   
\usepackage{hyperref}      
\usepackage{url}            
\usepackage{booktabs}      
\usepackage{amsfonts}      
\usepackage{nicefrac}      
\usepackage{microtype}     
\usepackage{xcolor}
\usepackage{pgfplots}
\pgfplotsset{compat=1.17}
\usepackage{wrapfig}
\title{SpatialVAM: Spatial-Aware Multi-View Video Diffusion as a Data-Efficient Robot Policy}

\author{
\textbf{Peiyan Li}$^{1,2,}$\thanks{Equal contribution},
\quad
\textbf{Yixiang Chen}$^{1,2,}$\footnotemark[1],
\quad
\textbf{Yuan Xu}$^{1,2}$,
\quad
\textbf{Jiabing Yang}$^{1,2}$,
\quad
\textbf{Xiangnan Wu}$^{1,2}$,  \\
\quad
\textbf{Jun Guo}$^{4}$,
\quad
\textbf{Nan Sun}$^{4}$,
\quad
\textbf{Long Qian}$^{5}$,
\quad
\textbf{Xinghang Li}$^{4}$,
\quad
\textbf{Xin Xiao}$^{6}$,
\quad
\textbf{Jing Liu}$^{3}$, \\
\quad
\textbf{Nianfeng Liu}$^{3}$,
\quad
\textbf{Tao Kong}$^{4,}$\thanks{Corresponding author},
\quad
\textbf{Yan Huang}$^{1,2,3,}$\footnotemark[2],
\quad
\textbf{Liang Wang}$^{1,2,}$,
\quad
\textbf{Tieniu Tan}$^{1,2,7}$ \\[8pt]
$^{1}$New Laboratory of Pattern Recognition (NLPR), \\Institute of Automation, Chinese Academy of Sciences \\
$^{2}$School of Artificial Intelligence, University of Chinese Academy of Sciences \\
$^{3}$FiveAges \quad
$^{4}$Tsinghua University \quad
$^{5}$Xi'an Jiaotong University \\
$^{6}$Wuhan University \quad
$^{7}$Nanjing University
}

\begin{document}

{
\hypersetup{linkbordercolor={1 1 1}}
\maketitle
}
\hypersetup{linkbordercolor={1 0 0}}

\vspace{-1cm}
\begin{center}
\url{https://spatialvam.github.io/}
\end{center}

\newcommand{\method}{SpatialVAM}
\input{paper/0-abstract}

\input{paper/1-introduction}
\input{paper/2-related_work}
\input{paper/3-method}
\input{paper/4-experiments}
\input{paper/5-conclusion}

\clearpage
\bibliographystyle{unsrtnat}
\bibliography{paper/reference}

\input{appendix}
\clearpage

\end{document}

%% file: paper/0-abstract.tex
\begin{abstract}
Robotic manipulation requires understanding both the 3D spatial structure of the environment and its temporal evolution, yet most existing policies neglect one or both aspects.
They often rely on 2D visual observations or backbones pretrained on static image--text pairs, which leads to high data requirements and limited comprehension of environment dynamics.
To address this, we introduce \method{}, \textbf{the first 3D Video Action Model} that simultaneously predict spatial-aware multi-view heatmap videos and RGB videos.
Our key insight is that this design naturally injects 3D information into video foundation models while aligning the representation format between video pretraining and action finetuning.
Extensive experiments demonstrate that \method{} enables \textbf{data-efficient, robust, generalizable, and interpretable} manipulation.
With only ten demonstration trajectories and no additional pretraining, \method{} handles challenging long-horizon and contact-rich tasks, generalizes to out-of-distribution settings, and predicts realistic future videos.
Evaluations on Meta-World (22\%$\uparrow$), RoboCasa (15\%$\uparrow$) and real-world robotic platforms (16\%$\uparrow$) show that \method{} consistently outperforms other video action models, vision language action models and 3D-based policies, establishing a new state-of-the-art in data-efficient multi-task manipulation.
\end{abstract}

%% file: paper/1-introduction.tex
\vspace{-1em}
\section{Introduction}
\begin{figure*}[t]
  \centering
  \includegraphics[width=1.0\textwidth]{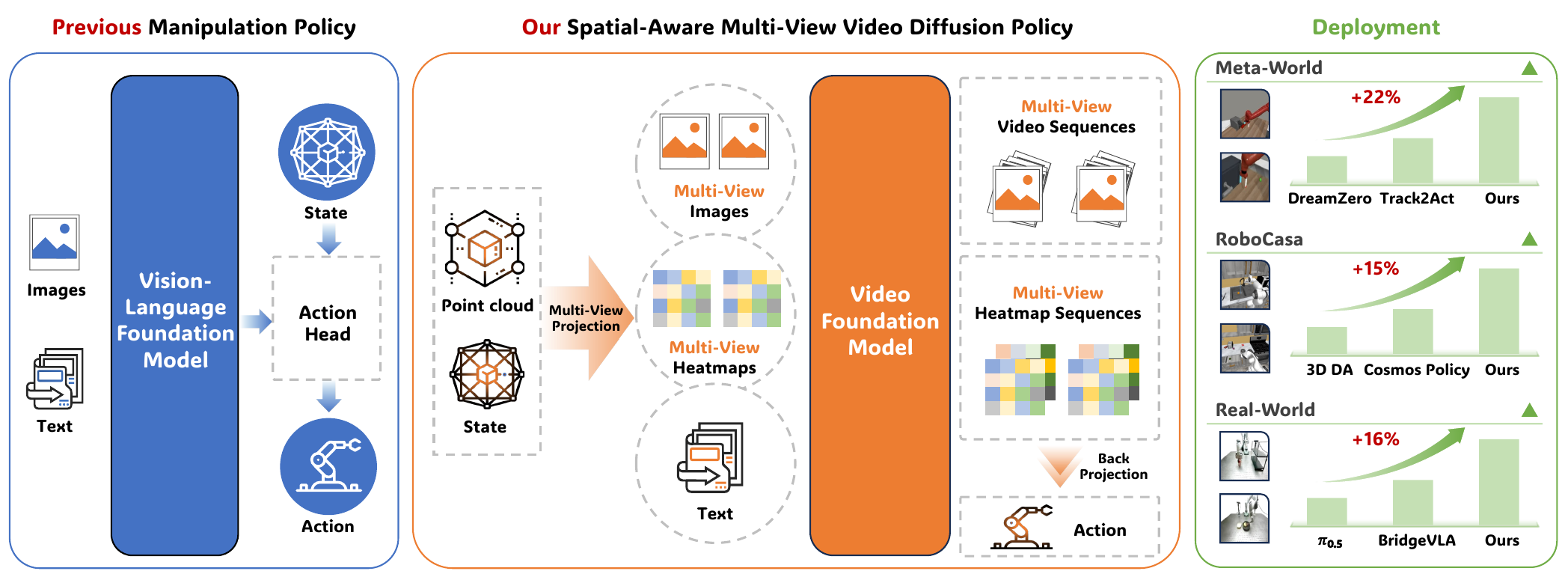}
\caption{
\textbf{Overview.} We introduce \method{}, a \textit{spatial-aware multi-view video diffusion policy} that jointly models spatial structure and temporal evolution. Compared to prior manipulation policies, our approach:
(1) processes \emph{spatial-aware multi-view images} rather than independent 2D views;
(2) represents robot actions as \emph{multi-view heatmaps}, aligning the action space with video pretraining representations;
(3) leverages a \emph{video foundation model}, instead of a vision--language backbone, to jointly model future RGB and heatmap sequences.
As a result, \method{} achieves state-of-the-art on Meta-World, RoboCasa and real-world benchmarks, outperforming vision language action models, video action models and 3D-based policies.
}
  \vspace{-0.6cm}
  \label{fig:teaser}
\end{figure*}

Recent years have witnessed substantial progress in robotics driven by large-scale representation learning, with Vision Language Action (VLA) models and Video Action Models (VAM) emerging as two representative paradigms~\cite{intelligence2025pi,intelligence2025pi_,black2410pi0,generalist2025gen0,zhai2025igniting,ye2026worldactionmodelszeroshot,kim2026cosmos}.
Despite their impressive performance across diverse tasks and embodiments, existing approaches still fall short of a unified understanding of the environment's \emph{3D spatio-temporal state}: they often lack either temporal modeling of how the world evolves or spatial modeling of the environment's 3D structure.

On the one hand, VLA-based methods commonly adopt backbones pretrained on static image--text pairs~\citep{beyer2024paligemma,wang2024qwen2,bai2025qwen2,liu2023visual}.
However, such pretraining does not capture how scenes evolve over time, limiting a policy's ability to anticipate future states.
This contrasts with insights from cognitive science, which suggest that effective action selection relies on anticipating how the environment will change in response to actions, rather than reacting to instantaneous observations~\citep{clark2013whatever}.
On the other hand, although VAMs inherit the predictive capabilities of video foundation models and naturally model future observations, they predominantly operate in 2D image space and lack explicit modeling of 3D geometric structure.
This creates an observation--action mismatch: observations are represented as 2D videos, while robot actions are executed in 3D physical space.
As a result, policies must compensate for this gap with large amounts of training data.

These observations motivate the following question:
\textbf{Can we design a 3D World-Action Model that incorporates 3D structural priors while explicitly modeling environmental dynamics?}
In this paper, we take a step toward this goal.
As illustrated in Fig.~\ref{fig:teaser}, we propose \method{}, a spatially aware multi-view video diffusion policy that \textbf{aligns the representation format of video pretraining with that of action finetuning}.
Specifically, \method{}:
(1) adopts spatially aware multi-view projections to implicitly encode 3D structure;
(2) transforms colored point clouds into multi-view RGB images and represents target positions as multi-view heatmaps using the same projection mechanism; and
(3) leverages a unified video foundation model to jointly predict future video frames and heatmap sequences, which are subsequently decoded into continuous action chunks.
By representing 3D information within this aligned video-based format, \method{} substantially reduces the gap between perception and control, leading to strong empirical performance.

Extensive experiments on Meta-World~\citep{yu2020meta}, RoboCasa~\citep{robocasa2024}, and real-world robotic platforms show that \method{} consistently outperforms state-of-the-art VAMs, VLA models, and 3D-based policies. These results highlight key strengths of \method{}: it is \textbf{data-efficient}, acquiring challenging skills from as few as 10 demonstrations without pretraining; \textbf{robust}, maintaining strong performance even with a single denoising step; \textbf{generalizable}, handling unseen backgrounds, object categories, heights, and lighting conditions; and \textbf{interpretable}, as predicted videos faithfully reflect executed actions, enabling users to visually assess action safety prior to deployment.

Our contributions are threefold:
(i) \textbf{Concept and insight:} we analyze key limitations of existing policies and introduce, to our knowledge, the first framework leveraging video foundation models for a 3D World-Action Model;
(ii) \textbf{Method:} we propose \method{}, a multi-view video diffusion policy that captures spatial structure and temporal dynamics and produces robot actions; and
(iii) \textbf{Experiments:} we perform extensive simulation and real-world evaluations to demonstrate \method{}'s advantages over existing manipulation policies.

%% file: paper/2-related_work.tex
\vspace{-1em}
\section{Related Work}
\vspace{-1em}
\subsection{Vision Language Action Models}
\vspace{-0.4em}
Vision Language Action (VLA) models have recently emerged as a dominant paradigm for robot  manipulation~\citep{intelligence2025pi,intelligence2025pi_,black2410pi0,generalist2025gen0,spiritai2026spiritv15,zhai2025igniting,cheang2025gr,cheang2024gr,li2025gr,qu2025eo,liu2025towards}. These models typically condition on 2D visual observations and language instructions, leverage large Vision--Language Models (VLMs) for representation learning, and decode continuous actions via flow matching~\citep{lipman2022flow}, diffusion~\citep{chi2025diffusion}, or fast tokenization~\citep{pertsch2025fast}.

While VLAs demonstrate strong generalization across tasks and embodiments~\citep{black2410pi0,intelligence2025pi_,intelligence2025pi}, they often require large-scale robot datasets and primarily emphasize action imitation, with limited explicit modeling of environment dynamics. Our approach addresses these limitations by implicitly encoding 3D-related information and integrating environmental dynamics modeling, ultimately achieving significantly higher data efficiency.
\vspace{-1em}
\subsection{Video Action Models}
\vspace{-0.3em}
Video action models have been regarded as a promising way for anticipatory control in robotic manipulation. Existing approaches can be broadly categorized into those that train video prediction and action generation separately, and those that train them jointly.

\textbf{Separate training.}
Many methods adopt a two-stage pipeline, first learning to predict future visual observations and then mapping visual representations to actions~\citep{hu2024video,pai2025mimic,du2023learning,yang2025roboenvision}. Variants differ in the intermediate representations used, including RGB videos~\citep{hu2024video,pai2025mimic}, human demonstration videos~\citep{bharadhwaj2024gen2act}, optical flow~\citep{ko2023learning}, and 2D point trajectories~\citep{bharadhwaj2024track2act}.

\textbf{Joint training.}
More recent work jointly predicts future videos and actions within a unified framework~\citep{wu2023unleashing,li2025unified,yang2025covar,Chen_2025_ICCV,zhu2025unified,guo2024prediction,li2026causal,ye2026worldactionmodelszeroshot,yuan2026fastwam,flowwam,ma2026dit4dit,bi2026motus}, enabling tighter perception–action coupling. Our method also follows this paradigm, but differs on implicitly incorporating 3D structural priors and alignment between action finetuning and video pretraining.
\vspace{-0.6em}
\subsection{3D Structural Priors for Manipulation}
Incorporating 3D structural priors into manipulation policies has been studied through several representations. Point-cloud-based methods directly encode 3D geometry for action prediction~\citep{gervet2023act3d,ze20243d,yang2025fp3,chen2023polarnet}, while other approaches use the knowledge in 3D foundation models as an implicit 3D proxy~\citep{qu2025spatialvla,li2025spatial}. More recently, multi-view image representations, combined with known camera transformations, have emerged as another effective form of 3D priors~\citep{goyal2024rvt,li2025bridgevla}.

Our work builds on this multi-view paradigm but differs in two aspects. First, existing approaches rely on vision-language models or task-specific networks, whereas \method{} employs a video foundation model to jointly capture spatial structure and temporal dynamics. Second, prior 3D-aware methods predict only key poses or waypoints, while \method{} directly predicts continuous actions and future observations, enabling more general manipulation skills.

%% file: paper/3-method.tex
\vspace{-0.3cm}
\section{Method}
\vspace{-0.1cm}
\begin{figure*}[t]
  \centering
  \includegraphics[width=1.0\textwidth]{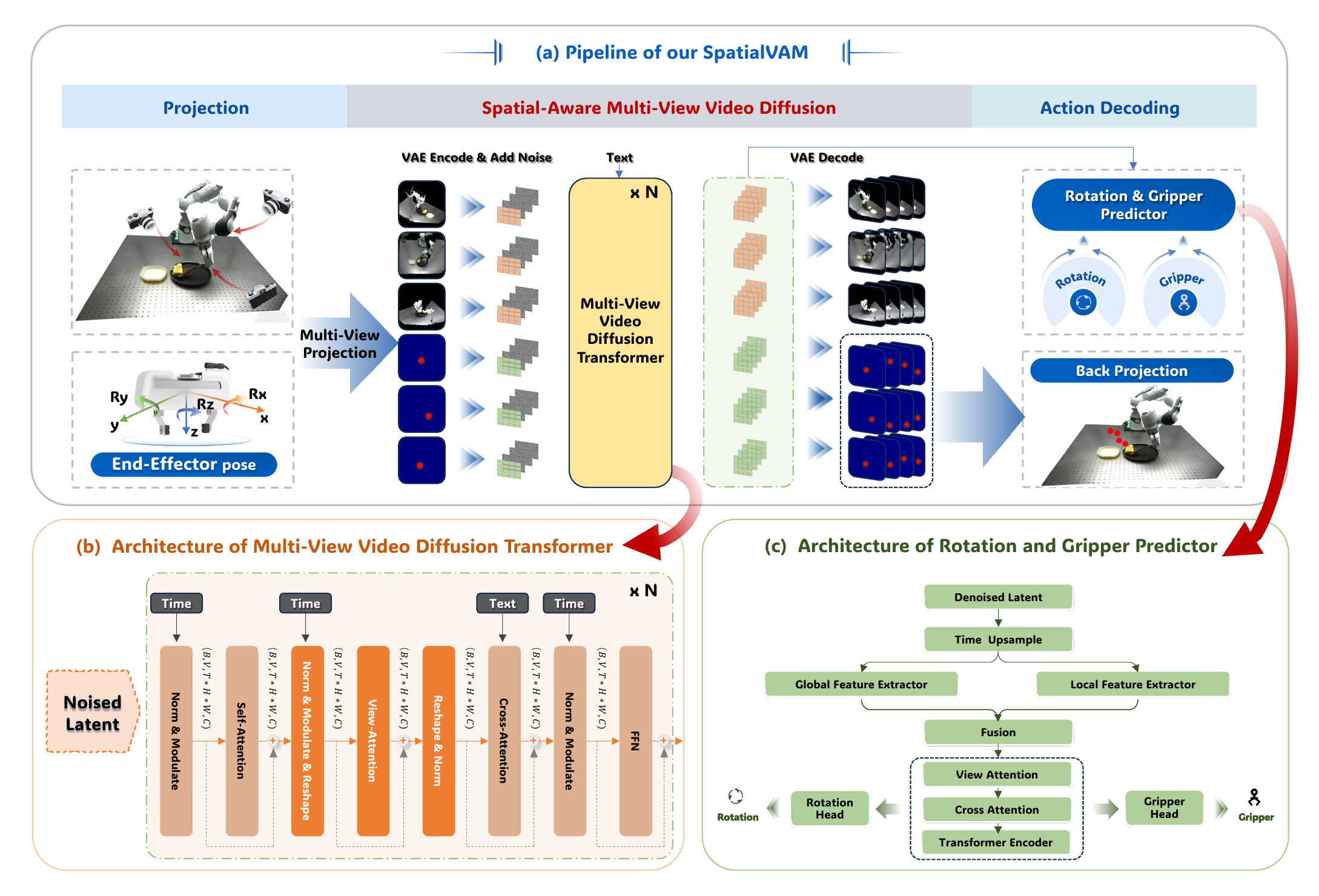}
\caption{
\textbf{Overview of \method{}'s pipeline}.
(a) Point clouds and the end-effector pose are projected into spatial-aware multi-view RGB images and heatmaps, which are encoded to jointly predict future multi-view RGB and heatmap videos via a video diffusion model. Predicted heatmaps are back-projected to recover 3D end-effector positions. (b) The multi-view video diffusion transformer augments a pretrained backbone with view-attention to enable cross-view interaction. (c) A lightweight action decoder predicts end-effector rotation and gripper states from the denoised video latents. Final action chunks combine the predicted positions, rotations, and gripper states.}
  \vspace{-0.2cm}
  \label{fig:method}
\vspace{-0.2cm}
\end{figure*}

As illustrated in Fig.~\ref{fig:method} (a), our \method{} consists of three main components, detailed below.
In Sec.~\ref{sec.method_proj}, we present the multi-view projection module that transforms colored point clouds into multi-view images and heatmap representations.
In Sec.~\ref{sec.method_diffusion}, we describe the multi-view video diffusion formulation and the corresponding model architecture.
In Sec.~\ref{sec.method_action_decode}, we detail how final actions are decoded from the predicted heatmaps and latent representations.
Finally, in Sec.~\ref{sec.method_training_inference}, we introduce the training and inference procedures of \method{}.

\vspace{-0.3cm}
\subsection{Multi-View Projection}
\label{sec.method_proj}
\vspace{-0.1cm}
\method{} takes point clouds and the robot end-effector pose as raw inputs.
Since redundant point cloud information can negatively affect subsequent model training~\citep{ze20243d}, we first crop the input point clouds to the workspace boundaries, defined as a cubic volume of $1\,\mathrm{m}^3$.

Instead of processing the cropped point clouds directly, we render them from three fixed virtual viewpoints using orthographic projection~\citep{goyal2024rvt,li2025bridgevla}.
The viewing directions need not be mutually orthogonal and can be flexibly chosen, as long as the target regions are observed as clearly as possible.
Although adaptively selecting projection views is a promising direction~\citep{chen2025verm}, we leave this exploration to future work.
The advantages of virtual-view rendering over directly using the physical-camera images are discussed in Appendix~\ref{app:why_virtual_views}.

For the end-effector pose, rather than encoding it with a separate module, we transform it into multiple heatmaps corresponding to the projection views.
After colorization, these heatmaps can be directly processed by the VAE encoder of the Video Foundation Model (VFM).
Our intuition is that leveraging off-the-shelf pretrained components reduces the representation gap and exploits prior knowledge.

To construct the heatmaps, we first identify the end-effector's corresponding pixel locations on each projection view.
Centered at these pixels, we generate Gaussian heatmaps $H_i^t$ with spatial truncation:
\begin{equation}
H_i^t(\mathbf{x})=
\begin{cases}
p_i^t(\mathbf{x}), & \text{if } p_i^t(\mathbf{x}) \geq \tau, \\
0, & \text{otherwise},
\end{cases}
\label{eq:single_object_probability_map}
\end{equation}
where $t$ denotes the timestep, $i \in \{1,2,...,n\}$ indexes the view, and $\mathbf{x}=(u,v)$ denotes the pixel location.
The untruncated Gaussian distribution is defined as
\begin{equation}
p_i^t(\mathbf{x})=\exp\left(-\frac{\|\mathbf{x}-\widehat{\mathbf{x}}_i^t\|^2}{2\sigma^2}\right),
\end{equation}
where $\widehat{\mathbf{x}}_i^t$ is the projected pixel location of the end effector at timestep $t$ on view $i$, $\sigma$ controls the spatial spread of the heatmap, and $\tau$ is a probability threshold.

At this stage, we obtain the current spatial-aware multi-view RGB images and the corresponding heatmap images, which serve as conditioning inputs for the multi-view video diffusion transformer to jointly predict future multi-view RGB videos and heatmap videos.

\subsection{Multi-View Video Diffusion}
\label{sec.method_diffusion}

Our multi-view video diffusion transformer is built upon Wan2.2~\citep{wan2025}, a 5B-parameter video foundation model pretrained on single-view videos.
For the multi-view setting, we augment each Diffusion Transformer (DiT) block~\citep{Peebles2022DiT} with a view-attention module~\citep{bai2024syncammaster}, as illustrated in Fig.~\ref{fig:method} (b).

During the forward pass, the multi-view RGB and heatmap images are encoded into latents by the pretrained VAE encoder. These latents are then concatenated along the view dimension, followed by patchifying and flattening across the spatial and temporal dimensions. This yields token sequences of shape $(B, V, T \times H \times W, C)$, where $B$, $V$, $T$, $H$, $W$, and $C$ are the batch size, number of views, time length, height, width, and channel count, respectively.

To accommodate the view-attention module, the token sequences are reshaped to $(B, T, V \!\times\! H \!\times\! W, C)$, enabling explicit interactions across views at each timestep.
After applying view attention, the tokens are reshaped back to their original layout and processed by the remaining transformer layers.
All other components follow the original Wan2.2 architecture to preserve the pretrained knowledge.

The transformer is trained to predict the added noise, enabling multi-step denoising to recover clean latent representations.
The resulting latents are decoded by the VAE decoder to generate future multi-view RGB and heatmap videos, which are subsequently used for action decoding. We provide the full formulation and additional details of the diffusion process in Appendix~\ref{app:diffusion_theory}.
\subsection{Action Decoding}
\label{sec.method_action_decode}

After obtaining the predicted heatmap sequences, we back-project the peak locations of the three heatmaps at each timestep into a 3D position in the workspace using known camera parameters.
Details of the projection and back-projection procedures are provided in Appendix~\ref{app.proj_back_proj}.
By back-projecting heatmaps across all predicted timesteps, we recover a continuous 3D end-effector trajectory.

For rotation and gripper prediction, as illustrated in Fig.~\ref{fig:method} (c), we take the denoised latent representations as input.
Since the latents are temporally compressed by the VAE, we first upsample them along the temporal dimension.
We then employ two convolutional networks to extract global features from the entire latent representation and local features centered around heatmap peak locations, respectively.
These features are fused and further aggregated along the view dimension.

The latent corresponding to the first frame serves as the conditioning latent.
We apply cross-attention between the predicted latents and the conditioning latent to incorporate conditioning information.
The resulting representations are encoded by a lightweight four-layer transformer, followed by two separate MLP heads to predict rotation and gripper actions.
Rotation and gripper states are discretized, and we predict their changes at each future timestep relative to the conditioning frame.

\vspace{-0.2cm}
\subsection{Training \& Inference}
\label{sec.method_training_inference}

Our \method{} consists of two trainable modules: (i) a 5B multi-view video diffusion transformer for predicting end-effector positions, and (ii) a lightweight 170M rotation \& gripper predictor. 
\paragraph{Training.}
During training, we apply SE(3) augmentations to the input point clouds and end-effector poses before projecting them into multi-view RGB images and heatmaps.
These representations are encoded by a shared VAE encoder and concatenated along the view dimension.

To train the multi-view video diffusion transformer, Gaussian noise is added to future-frame latents, with the noise magnitude determined by randomly sampled diffusion timesteps. The model is then trained to predict the noise using an MSE loss. Since we predict both videos and heatmap sequences, the total training loss for the multi-view video diffusion transformer consists of two components:
\begin{equation}
L_{diff} = \lambda L_{vid} + (1 - \lambda) L_{heat},
\end{equation}
where \( L_{vid} \) is the diffusion loss for the video sequences, \( L_{heat} \) is the diffusion loss for the heatmap sequences, and \( \lambda \) is the weight applied to the video diffusion loss (see Sec.~\ref{exp:robustness} for an analysis of key parameters' robustness).
To reduce computational cost, we adopt LoRA fine-tuning. While we also experimented with full fine-tuning (see Sec.~\ref{exp:ablation}), we observed no performance gains.

For training the rotation and gripper predictor, we use ground-truth video latents and heatmap latents as inputs. To enhance robustness, we inject a small amount of random noise into the ground-truth latents. The training targets include the end-effector's rotation and gripper state. Rotation is represented using Euler angles, discretized into 72 bins, while the gripper state is modeled as a binary variable. The module is trained using cross-entropy loss, formulated as:
\begin{equation}
L_{pred} = L_{rol} + L_{pit} + L_{yaw} + L_{gri},
\end{equation}
where $L_{rol}$, $L_{pit}$, $L_{yaw}$ are Euler-angle losses and $L_{gri}$ is the gripper loss.
\vspace{-0.3cm}
\paragraph{Inference.}
During inference, we project the point cloud and end-effector pose at the current timestep into multi-view RGB images and heatmaps, which are encoded to obtain the conditioning latent.
After obtaining the predicted clean latents, we decode them through two parallel branches.
One branch decodes the latents into multi-view heatmap sequences, which are back-projected to recover 3D position predictions.
The other branch feeds the latents into the rotation \& gripper predictor to estimate rotation and gripper states.
By combining the predicted positions, rotations, and gripper states, we obtain action chunks that are subsequently executed by the robot controller.
Additional implementation details and hyperparameters are provided in Appendix~\ref{app:details on train and infer}.

%% file: paper/4-experiments.tex
\section{Experiments}
We conduct extensive experiments in both simulation and real-world settings to evaluate the effectiveness of \method{}. Our evaluation is designed to address several key questions: whether \method{} outperforms other Video Action Models (\textbf{Q1}), 3D-based policies (\textbf{Q2}), and Vision-Language-Action models (\textbf{Q3}); whether it generalizes to unseen scenarios (\textbf{Q4}); remains robust to changes in deployment conditions and modeling parameters (\textbf{Q5}); benefits from each of its architectural components (\textbf{Q6}); and produces videos that faithfully reflect executed actions to facilitate safer deployment (\textbf{Q7}).
\vspace{-0.3cm}
\subsection{Meta-World Experiments}
\label{exp:meta_world}
\vspace{-0.1cm}
\paragraph{Setup and Baselines}
Meta-World~\citep{yu2020meta} is a standardized tabletop manipulation benchmark with a simulated Sawyer arm. We train \method{} on seven tasks with only five expert demonstrations per task (35 total, a low-data regime), and running 25 random-initialization trials per task with a 600-step success horizon for evaluation.
For baselines, we compare our method against standard behavioral-cloning and video-prediction-based baselines. BC-Scratch and BC-R3M~\citep{nair2022r3m} are multi-task behavioral-cloning baselines built on a ResNet-18 visual encoder and a CLIP text encoder, where BC-R3M is initialized with R3M-pretrained weights. Diffusion Policy (DP)~\citep{chi2025diffusion} models action sequences via a diffusion-based formulation. Our video-prediction-based baselines include UniPi~\citep{du2023learning}, AVDC~\citep{ko2023learning}, Track2Act~\citep{bharadhwaj2024track2act}, and DreamZero~\citep{ye2026worldactionmodelszeroshot}. UniPi, AVDC, and Track2Act predict future RGB frames, optical flow, and 2D point trajectories, respectively, and then infer actions from these predicted visual representations. DreamZero is a state-of-the-art video action model that also adopts Wan~\citep{wan2025} as its backbone, jointly predicting future RGB videos and the corresponding actions. Additional details on these baselines are provided in Appendix~\ref{app:sim_baselines}.
\paragraph{Results.}
As shown in Tab.~\ref{tab:metaworld_results}, standard behavioral cloning methods perform poorly in the low-data regime, achieving only 26.2\%, 35.4\%, and 37.7\% success rates for BC-Scratch, BC-R3M, and DP, respectively.
In contrast, video-prediction-based methods yield substantially stronger performance, with AVDC, Track2Act, and DreamZero reaching 58.9\%, 67.4\%, and 61.1\%, respectively.
\method{} outperforms all baselines on five out of seven tasks and achieves the best overall average success rate of 89.1\% using only 5 demonstrations per task.
These results suggest three main conclusions. First, video prediction is highly beneficial for data-efficient manipulation learning. Second, our multi-view video diffusion formulation built on a video foundation model is more effective than prior video-prediction-based approaches. Third, 3D-aware multi-view prediction plays a critical role: although DreamZero uses the same video foundation model as our method, it performs substantially worse. Together, these findings provide a direct answer to \textbf{Q1}.
\input{tables/meta_world_results}
\begin{wrapfigure}[12]{r}{0.36\textwidth}
\setlength{\intextsep}{0pt}%
\vspace{-0.4em}

\centering
\begin{tikzpicture}
\begin{axis}[
    ybar,
    bar shift=0pt,
    width=\linewidth,
    height=3.8cm,
    ymin=0, ymax=55,
    ytick={0,10,20,30,40,50},
    bar width=18pt,
    enlarge x limits=0.18,
    ylabel={Avg. Succ. Rate (\%)},
    ylabel style={font=\scriptsize, yshift=-3pt},
    symbolic x coords={dda,cosmos,ours},
    xtick=\empty,
    yticklabel style={font=\scriptsize},
    nodes near coords={\pgfmathprintnumber[fixed, precision=1]{\pgfplotspointmeta}},
    nodes near coords style={font=\scriptsize},
    axis lines*=left,
    tick style={draw=none},
    clip=false,
]
\addplot[draw=black, fill=orange!65] coordinates {(dda,7.6)};
\addplot[draw=black, fill=teal!55]   coordinates {(cosmos,20.0)};
\addplot[draw=black, fill=blue!65]   coordinates {(ours,35.8)};

\draw[fill=orange!65, draw=black!70] ([xshift=-0.13cm, yshift=0.70cm] axis cs:dda,55)    rectangle ++(0.26cm, 0.11cm);
\draw[fill=teal!55,   draw=black!70] ([xshift=-0.13cm, yshift=0.70cm] axis cs:cosmos,55) rectangle ++(0.26cm, 0.11cm);
\draw[fill=blue!65,   draw=black!70] ([xshift=-0.13cm, yshift=0.70cm] axis cs:ours,55)   rectangle ++(0.26cm, 0.11cm);

\node[anchor=north, font=\scriptsize, inner sep=1pt]                at ([yshift=0.67cm] axis cs:dda,55)    {3D DA};
\node[anchor=north, font=\scriptsize, inner sep=1pt, align=center]  at ([yshift=0.67cm] axis cs:cosmos,55) {Cosmos\\Policy};
\node[anchor=north, font=\scriptsize, inner sep=1pt]                at ([yshift=0.67cm] axis cs:ours,55)   {SpatialVAM};
\end{axis}
\end{tikzpicture}
\vspace{0.8em}
\caption{RoboCasa results. 10 demos/task for training and 50 rollouts for evaluation.}
\label{fig:robocasa_results}
\end{wrapfigure}
\vspace{-0.9cm}
\subsection{RoboCasa Experiments}
\label{exp:robocasa}
To further evaluate the effectiveness of \method{}, we conduct experiments on ten kitchen manipulation tasks from RoboCasa~\cite{robocasa2024}, a simulation benchmark built on the MuJoCo stack. RoboCasa allows users to procedurally generate diverse kitchen environments with varying styles, furniture configurations, and object instances. It provides the Franka Emika Panda robot as the default manipulation embodiment, with the end-effector controlled in 6-DoF Cartesian delta space.

We compare \method{} against two baselines on this benchmark. First, Cosmos Policy~\cite{kim2026cosmos}, a representative Video Action Model that formulates both action and value prediction as video diffusion by repeating them as images. While this design helps reduce the gap between video pretraining and action finetuning, it does not incorporate any 3D structural priors. Second, 3D Diffuser Actor~\cite{3d_diffuser_actor}, which integrates 3D input priors with diffusion-based action modeling. Although it utilizes 3D information, it does not leverage a video foundation model and lacks the ability to predict future environmental evolution.

For each task, we train with only 10 demonstrations and evaluate performance over 50 rollout trials. As shown in Fig.~\ref{fig:robocasa_results}, \method{} outperforms both baselines by a substantial margin, further demonstrating its effectiveness and exceptional data efficiency. These results provide additional evidence in support of \textbf{Q1} and \textbf{Q2}.
\subsection{Real-World Experiments}
\label{exp:real_world}
\begin{figure*}[t]
  \centering
  \includegraphics[width=\textwidth]{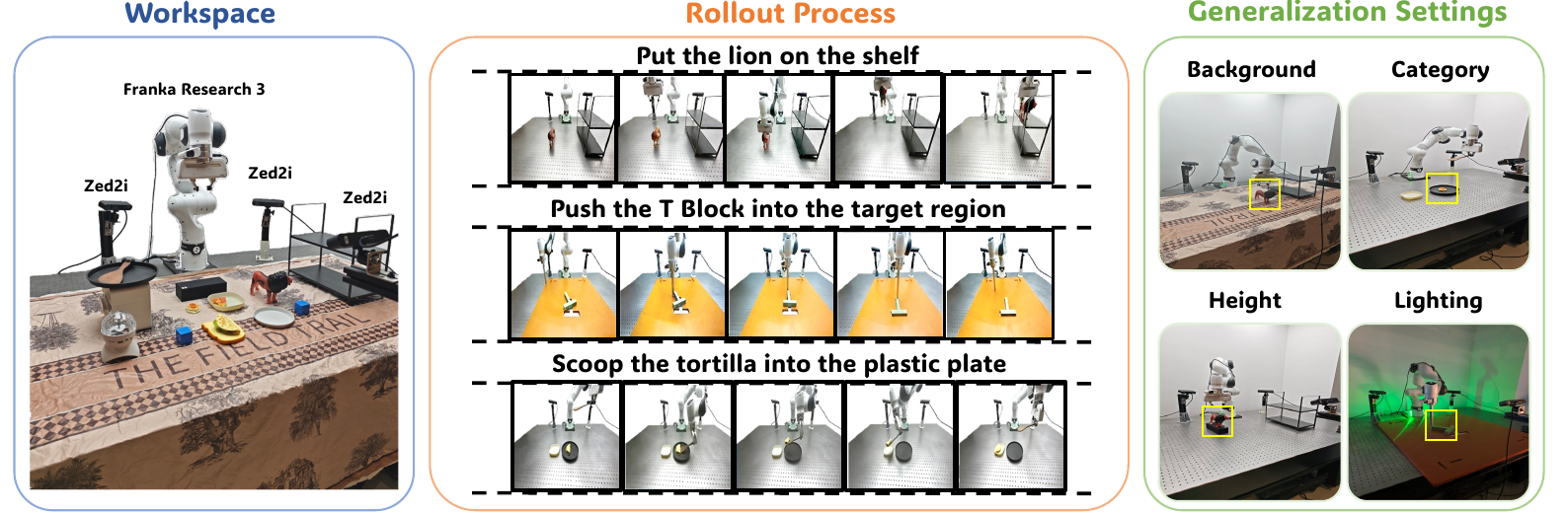}
\caption{\textbf{Real-world experimental setup and tasks.} We evaluate \method{} on three manipulation tasks using a Franka Research 3 robot with three ZED2i cameras. We further assess generalization under variations in background, object height, lighting, and object category.}
  \vspace{-0.5cm}
  \label{fig:real_setup}
\end{figure*}
\paragraph{Setup.}
As shown in Fig.~\ref{fig:real_setup}, our real-world platform consists of a Franka Research 3 arm with a standard parallel-jaw gripper, observed by three static ZED~2i depth cameras. We calibrate the cameras with respect to the robot base frame and transform their RGB-D observations into colored point clouds as input to \method{}.

We evaluate \method{} on three base manipulation tasks: (i) a simple pick-and-place task (\textit{Put Lion}), (ii) a complex pushing T-shaped block task (\textit{Push-T}), and (iii) a contact-rich scooping task (\textit{Scoop Tortilla}).
To assess generalization, we introduce variations to the original task settings, resulting in four unseen tasks:
(1) \textbf{Put-B}: a visually distinct cloth is placed on the table in the \textit{Put Lion} task;
(2) \textbf{Put-H}: the target lion is placed on a $5.5\,\mathrm{cm}$-high box in the \textit{Put Lion} task;
(3) \textbf{Push-L}: the ambient lighting is turned off in the \textit{Push-T} task;
(4) \textbf{Scoop-C}: the tortilla is replaced with plastic noodles, representing an unseen object category.
Fig.~\ref{fig:real_setup} provides visual illustrations of all tasks.
 
We collect about ten expert trajectories per task using a SpaceMouse for teleoperation and \texttt{frankapy}~\citep{zhang2020modular} as the control interface. Each task is evaluated for ten trials. For fair comparison, we photograph each test scene and manually align the environment setup across all evaluated methods, ensuring that every method encounters identical initial scene configurations.  
      
\paragraph{Baselines.}
We compare our method with representative approaches spanning three dominant paradigms—3D-based policies, video-prediction–based methods, and VLA models. A brief overview is provided here, with a more detailed description in Appendix~\ref{app:real_world_baselines}.
\begin{itemize}[leftmargin=1.5em, itemsep=1pt, topsep=2pt]
    \item \textbf{DP3}~\citep{ze20243d}: a state-of-the-art 3D visuomotor policy that encodes point clouds using an MLP and predicts actions via a diffusion-based policy head.

    \item $\boldsymbol{\pi_{0.5}}$~\citep{intelligence2025pi_}: a large-scale vision-language-action model pretrained on diverse data and decoding actions using flow matching.

    \item \textbf{UVA}~\citep{li2025unified}: a video-prediction–based policy that jointly predicts future videos and actions using unified representations and dual diffusion heads.

    \item \textbf{BridgeVLA}~\citep{li2025bridgevla}: a 3D vision-language-action model that projects point clouds into 3D-aware multi-view images and predicts heatmaps for action decoding.
\end{itemize}
\vspace{-1em}
\paragraph{Results.}
The quantitative results are reported in Tab.~\ref{tab:real_results}.
\input{tables/real_result}
Under the extremely limited data regime, only BridgeVLA~\citep{li2025bridgevla} and \method{} achieve reasonable success rates on the basic tasks.
DP3 overfits, owing to its simple MLP point-cloud encoder (addressing \textbf{Q2});
$\pi_{0.5}$ exhibits limited spatial generalization beyond the training distribution (\textbf{Q3});
UVA exhibits qualitatively correct action tendencies but insufficient execution precision (\textbf{Q1});
BridgeVLA, although built on the same multi-view heatmap paradigm as \method{}, predicts only single key poses and therefore fails on the continuous-motion \textit{Push T} task and the contact-rich \textit{Scoop Tortilla} task.
In contrast, \method{} leverages a video foundation model to predict continuous multi-view heatmap and RGB sequences, enabling it to solve all three basic tasks; detailed failure analyses of \method{} and the baselines are provided in Appendix~\ref{app:real_world_detail}.
On unseen tasks, \method{} generalizes strongly, achieving the best performance on the \textit{Push-L} and \textit{Scoop-C} settings (\textbf{Q4}); it slightly underperforms BridgeVLA on \textit{Put-B} and \textit{Put-H}, a gap we attribute to BridgeVLA's vision--language backbone being pretrained on additional image--text pairs, which better captures such appearance variations. We think scaling up our video pretraining could analogously improve the generalization ability.

\subsection{Robustness Analysis}
\label{exp:robustness}
We evaluate the robustness of \method{} from two complementary perspectives: deployment robustness to changes in camera coverage and calibration, and modeling robustness to key hyperparameters and inference settings.

\paragraph{Deployment robustness.}
Our default setup fuses observations from three physical cameras to obtain a more complete point cloud. To examine whether multiple cameras are essential, we retain only one camera, which substantially reduces the observed point-cloud coverage. Even with this incomplete observation, using the original virtual projection views only moderately reduces the overall success rate from 57.1\% to 50.0\%. This decline stems largely from a view mismatch: the projection views selected for the three-camera point cloud no longer expose the most informative regions visible to the remaining camera. After adapting the virtual views to better cover the observable workspace, the success rate recovers to 60.0\%, comparable to the three-camera result. The projected observations under these three settings are visualized in Fig.~\ref{fig:camera_view_visualization}, highlighting how the adapted views better expose the observable workspace from the single-camera point cloud. These results show that \method{} does not fundamentally rely on multiple physical cameras and remains effective with incomplete point clouds when its virtual projection views are aligned with the observable workspace.

To evaluate robustness to substantial calibration errors, we conduct a deliberately challenging stress test by randomly perturbing the camera extrinsics by $20\,\mathrm{mm}$ in translation and $2^\circ$ in rotation. When training uses the original calibration but evaluation uses the perturbed calibration, \method{} retains a 51.4\% success rate, compared with 57.1\% under the original calibration. When training and evaluation share the same perturbation, the success rate reaches 54.3\%, approaching the original-calibration result. We attribute this robustness to our projection and heatmap representations: calibration errors become localized spatial displacements after rendering, while the spatially smooth heatmaps reduce sensitivity to exact coordinates and allow consistent calibration biases to be absorbed during training. Thus, highly accurate calibration is not a strict requirement for \method{}. Full protocols and per-task results for both deployment robustness studies are provided in Appendix~\ref{app:camera_calibration_robustness}.

\paragraph{Modeling robustness.}
Beyond deployment conditions, we examine whether the strong performance of \method{} depends on careful tuning of its modeling and inference parameters. On Meta-World, we vary three key parameters: the RGB loss weight $\lambda$, heatmap standard deviation $\sigma$, and inference denoising steps $N$. The results for $\lambda$ and $\sigma$ are summarized in Appendix~\ref{app:robustness_analysis}. As shown, varying $\lambda$ by 80\% results in only a 3.3\% change in the success rate, while varying $\sigma$ by 133\% leads to a mere 2.5\% change. These findings show that the performance gains of \method{} do not rely on narrowly tuned hyperparameters.

\begin{figure*}[t]
  \centering
  \begin{minipage}[t]{0.48\textwidth}
    \centering
    \includegraphics[width=\linewidth]{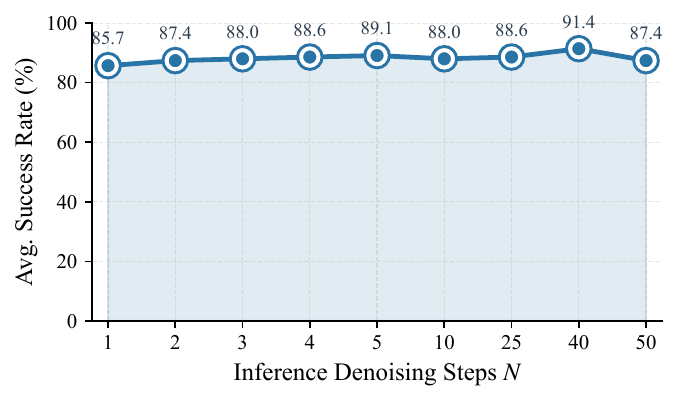}
  \end{minipage}
  \hfill
  \begin{minipage}[t]{0.48\textwidth}
    \centering
    \includegraphics[width=\linewidth]{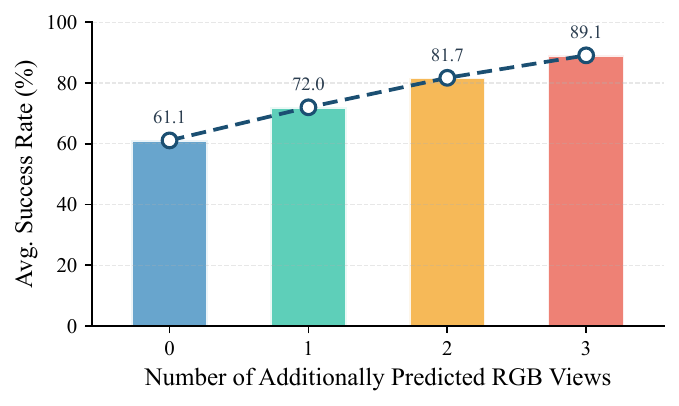}
  \end{minipage}
  \vspace{-0.25cm}
  \caption{\textbf{Robustness and ablation on Meta-World.}
  \textbf{Left:} Avg. success rate vs. inference denoising steps $N$; \method{} stays strong even with $N{=}1$.
  \textbf{Right:} Avg. success rate vs. number of additionally predicted RGB views; predicting more views yields monotonic improvements.}
  \vspace{-0.3cm}
  \label{fig:robustness_and_views}
\end{figure*}

We next examine the inference denoising steps $N$, varying it from 1 to 50. The results are presented in Fig.~\ref{fig:robustness_and_views}~(left). Surprisingly, in contrast to typical video diffusion processes~\citep{yang2024cogvideox,wan2025}, which require around 50 denoising steps, \method{} achieves a comparable success rate with just a single denoising step, substantially reducing inference cost. We believe the main reason for this is that heatmaps have relatively simple distribution modes and lack the high-frequency details present in regular images. Furthermore, our action prediction relies solely on the peak locations of the heatmaps, meaning that the overall quality of the heatmaps is less critical, and thus fewer denoising steps are required (qualitative visualizations under different $N$ are provided in Appendix~\ref{app:vis_video}). To balance performance with computational efficiency, we recommend setting the denoising step to 5, which allows for 5Hz inference frequency on a single NVIDIA A100 GPU server (see Appendix~\ref{app:details on train and infer}).

\subsection{Ablation Studies}
\label{exp:ablation}
We conduct extensive ablation studies on the Meta-World benchmark to validate the key design choices in \method{}; results are summarized in Tab.~\ref{tab:ablation}, ordered by success rate. Model~\#2 applies full fine-tuning instead of LoRA and reaches 87.4\%---comparable to LoRA but with substantially higher computational and memory cost---so we adopt LoRA throughout. Model~\#3 concatenates heatmaps and RGB frames along the channel dimension (with an additional convolutional adapter to match the DiT's input channels); the resulting information bottleneck drops performance to 81.1\%, confirming that view-dimension concatenation better preserves multi-view information. Model~\#4 bypasses heatmap back-projection and regresses 3D translation directly from the denoised latents with a transformer, and the success rate falls markedly to 76.6\%, validating that heatmap-peak back-projection is substantially more effective than latent-space regression for translation decoding. Model~\#5 removes the pretrained video-foundation-model weights and almost completely fails (4.6\%), as the model cannot even fit the training set, underscoring the essential role of large-scale video pretraining for data-efficient manipulation. Finally, Fig.~\ref{fig:robustness_and_views}~(right) varies the number of additionally predicted RGB views from 0 to 3; success rate grows monotonically from 61.1\% to 89.1\%, showing that (i) jointly predicting RGB and heatmap videos is markedly better than predicting heatmaps alone, and (ii) predicting more views yields a more complete spatio-temporal understanding and better downstream control. Together these results validate our design choices, answering \textbf{Q6}.

\vspace{-0.4em}
\begin{table*}[t]
  \centering
  \begin{minipage}[t]{0.64\textwidth}
    \captionof{table}{\textbf{Ablation on \method{} design choices.}
    \emph{View Concat}: concatenate heatmap and RGB latents along the view dimension.
      \emph{Init W}: use the pretrained video-foundation-model weights.
      \emph{Trans. Decode}: heatmap-peak back-projection (Heatmap) vs.\ transformer regression from latents (TF-Regress).
      \emph{LoRA}: adapt the pretrained transformer via LoRA rather than updating all its weights.
      Model~\#1 (bold) is our default; each subsequent row modifies a single design choice.}
    \label{tab:ablation}
    \vspace{4pt}
    \centering
    \footnotesize
    \renewcommand{\arraystretch}{1.08}
    \setlength{\tabcolsep}{4.5pt}
    \resizebox{\linewidth}{!}{%
    \begin{tabular}{c|cccc|c}
      \toprule
      \textbf{\#} & \textbf{View Concat} & \textbf{Init W} & \textbf{LoRA} & \textbf{Trans. Decode} & \textbf{Avg (\%)} \\
      \midrule
      1 & $\checkmark$ & $\checkmark$ & $\checkmark$ & Heatmap      & \textbf{89.1} \\
      2 & $\checkmark$ & $\checkmark$ & -            & Heatmap      & 87.4 \\
      3 & -            & $\checkmark$ & $\checkmark$ & Heatmap      & 81.1 \\
      4 & $\checkmark$ & $\checkmark$ & $\checkmark$ & TF-Regress   & 76.6 \\
      5 & $\checkmark$ & -            & $\checkmark$ & Heatmap      & 4.6  \\
      \bottomrule
    \end{tabular}}
  \end{minipage}%
  \hfill
  \begin{minipage}[t]{0.32\textwidth}
    \vspace*{0pt}
    \centering
    \vspace*{2pt}
    \includegraphics[width=0.78\linewidth]{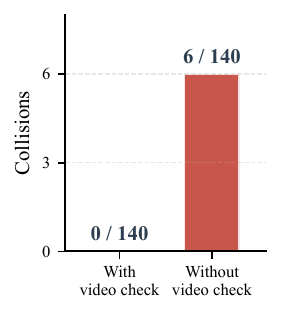}
    \captionof{figure}{\textbf{Collision events with and without video-based action checking} across 140 rollouts from four evaluators; video inspection reduces unsafe behaviors (discussed in Sec.~\ref{exp:safer_deployment}).}
    \label{fig:human_intervention}
  \end{minipage}
  \vspace{-0.3cm}
\end{table*}

\subsection{Safer Deployment and Enhanced Explainability}
\label{exp:safer_deployment}
\vspace{-0.1cm}
Deploying manipulation policies safely is challenging: the safety of a predicted action sequence is hard to judge from its raw numerical form, and reliable verification often requires costly and potentially unsafe physical execution. In contrast, \method{} produces realistic, temporally consistent multi-view videos and heatmaps that users can visually inspect before execution, providing an interpretable safety check. To quantify this benefit and answer \textbf{Q7}, we conducted a user study with four evaluators, each performing 35 rollouts (140 in total), rerunning any rollout that looked unsafe (e.g., potential collisions); Fig.~\ref{fig:human_intervention} shows that this video-based inspection significantly reduces collisions. Qualitative examples of predicted videos and heatmaps are provided in Appendix~\ref{app:vis_video}.

%% file: tables/meta_world_results.tex
\begin{table*}[t]
\centering
\caption{
\textbf{Success rates on seven Meta-World tasks under a low-data regime (5 demonstrations per task).}
Each entry reports the number of successful rollouts out of 25 trials.
\method{} achieves the highest average success rate, outperforming all baselines.
}
\label{tab:metaworld_results}

\footnotesize

\setlength{\tabcolsep}{3pt}
\renewcommand{\arraystretch}{1.0}

\begin{tabularx}{\textwidth}{
>{\raggedright\arraybackslash}l
*{7}{>{\centering\arraybackslash}X}
>{\centering\arraybackslash}p{0.08\textwidth} 
}
\toprule
\multirow{2}{*}{Method}
& \multicolumn{7}{c}{Meta-World Tasks}
& \multirow{2}{*}{\makecell{Avg.\\Succ. (\%) $\uparrow$}} \\
\cmidrule(lr){2-8}
& \mbox{D-Open} & \mbox{D-Close} & Btn & \mbox{Btn-Top} & \mbox{Fct-Cls} & \mbox{Fct-Open} & Handle & \\
\midrule
UniPi\textsuperscript{\protect\citep{du2023learning}}
& 0/25 & 9/25 & 3/25 & 0/25 & 1/25 & 3/25 & 4/25
& \num{11.4} \\
BC-Scratch\textsuperscript{\protect\citep{nair2022r3m}}
& 6/25 & 9/25 & 9/25 & 3/25 & 5/25 & 5/25 & 9/25
& \num{26.2} \\
BC-R3M\textsuperscript{\protect\citep{nair2022r3m}}
& 1/25 & 15/25 & 9/25 & 1/25 & 6/25 & 17/25 & 13/25
& \num{35.4} \\
DP\textsuperscript{\protect\citep{chi2025diffusion}}
& 12/25 & 12/25 & 10/25 & 5/25 & 6/25 & 15/25 & 6/25
& \num{37.7} \\
AVDC\textsuperscript{\protect\citep{ko2023learning}}
& 18/25 & 23/25 & 15/25 & 6/25 & 14/25 & 6/25 & 21/25
& \num{58.9} \\ DreamZero\textsuperscript{\protect\citep{ye2026worldactionmodelszeroshot}}
& 0/25 & 11/25 & 23/25 & 3/25 & \textbf{20/25} & \textbf{25/25} & \textbf{25/25}
& \num{61.1} \\
Track2Act\textsuperscript{\protect\citep{bharadhwaj2024track2act}}
& 22/25 & 19/25 & 14/25 & 10/25 & 12/25 & 22/25 & 19/25
& \num{67.4} \\
\midrule
\textbf{\method{} (Ours)}
& \textbf{25/25} & \textbf{25/25} & \textbf{25/25} & \textbf{24/25} & 8/25 & 24/25 & \textbf{25/25}
& \textbf{\num{89.1}} \\
\bottomrule
\end{tabularx}

\vspace{-0.5cm}
\end{table*}

%% file: tables/real_result.tex
\begin{table*}[t]
\centering
\caption{\textbf{Real-world manipulation results under limited demonstrations.} Success rates over 10 trials per task on three basic and four unseen tasks. All methods use 10 expert trajectories.}   \label{tab:real_results}

\footnotesize 
\setlength{\tabcolsep}{2.0pt}
\renewcommand{\arraystretch}{1.2}

\begin{tabularx}{\textwidth}{
>{\raggedright\arraybackslash}p{0.18\textwidth}
*{7}{>{\centering\arraybackslash}X}
>{\centering\arraybackslash}p{0.08\textwidth}
}
\toprule
\multirow{2}{*}[-0.5ex]{Method} 
& \multicolumn{3}{c}{Basic Tasks}
& \multicolumn{4}{c}{Unseen Tasks}
& \multirow{2}{*}[-0.5ex]{\makecell{Avg.\\ Succ. (\%)}} \\ 
\cmidrule(lr){2-4}\cmidrule(lr){5-8}
& Put Lion
& Push-T
& Scoop Tort. 
& Put-B
& Put-H
& Push-L
& Scoop-C
& \\
\midrule
DP3\textsuperscript{\cite{ze20243d}}
& 0/10 & 0/10 & 0/10
& 0/10 & 0/10 & 0/10 & 0/10
& 0.00 \\
$\pi_{0.5}$\textsuperscript{\cite{intelligence2025pi_}}
& 1/10 & 0/10 & 0/10
& 0/10 & 0/10 & 0/10 & 0/10
& 1.40 \\
UVA\textsuperscript{\cite{li2025unified}}
& 2/10 & 0/10 & 0/10
& 1/10 & 1/10 & 0/10 & 0/10
& 5.70 \\
BridgeVLA\textsuperscript{\cite{li2025bridgevla}}
& 9/10 & 0/10 & 4/10
& \textbf{8/10} & \textbf{7/10} & 0/10 & 1/10
& 41.42 \\
\midrule
\textbf{SpatialVAM (Ours)} 
& \textbf{10/10} & \textbf{4/10} & \textbf{7/10}
& 5/10 & 6/10 & \textbf{3/10} & \textbf{5/10}
& \textbf{57.10} \\
\bottomrule
\end{tabularx}

\vspace{-0.6em}
\end{table*}

%% file: paper/5-conclusion.tex
\vspace{-0.2cm}
\section{Conclusion and Future Work}

In this paper, we introduced \method{}, a spatial-aware multi-view video diffusion policy that incorporates 3D structural priors into a video foundation model backbone. Our experiments demonstrate that \method{} is data-efficient, robust, generalizable, and interpretable. Nevertheless, our current real-world pipeline assumes at least one fixed RGB-D camera with approximately known calibration, adding deployment complexity. Inference also remains relatively slow: generating a 24-frame action chunk takes approximately 4.6\,s on an NVIDIA A100 GPU, making the method unsuitable for certain high-frequency dexterous tasks. In future work, we plan to integrate diffusion acceleration techniques, such as distillation~\cite{akbari2026flash}, quantization~\cite{li2026deltaquant}, asynchronous execution~\cite{black2026real}, and asymmetric denoising~\cite{li2026efficient}, to enable real-time control in practical robotic applications.

%% file: appendix.tex
\newpage
\appendix
\onecolumn
\begin{center}
{\LARGE \textbf{Spatial-Aware Multi-View Video
Diffusion \\ $\;$ \\ ————Appendix————}}
\end{center}

\section{Multi-View Video Diffusion}
\label{app:diffusion_theory}

In this section, we provide a detailed description of the multi-view video diffusion process used in \method{}, which is based on the Wan2.2 framework~\cite{wan2025}. This process consists of three main components: the 3D VAE encoder, the Diffusion Transformer (DiT) block, and the 3D VAE decoder. We will explain about these components and how they integrate within our pipeline.

\subsection{3D VAE Encoder}
\label{app:vae_encoder}

The first step in the video diffusion process is encoding the input RGB image sequences and heatmap sequences (which should first be colorized to have 3 RGB channels) into a latent space using a 3D Variational Autoencoder (VAE). Both the input RGB sequence and the heatmap sequence are treated as a video sequence $V \in \mathbb{R}^{(1+T) \times H \times W \times 3}$, where $(1+T)$ is the total number of frames, $H$ and $W$ are the spatial dimensions of each frame, and 3 denotes the RGB color channels.

The 3D VAE performs both spatial and temporal compression to reduce the dimensionality of the input video.

\textbf{Spatial Compression:}
The spatial compression is achieved by applying a series of 3D convolutions, which reduce the spatial resolution of the input video from $H \times W$ to $H/8 \times W/8$, while expanding the number of channels to $C=16$. This allows the model to capture important spatial features in a lower-dimensional representation.

\textbf{Temporal Compression:}
In addition to spatial compression, the 3D VAE also reduces the temporal resolution by applying 3D convolutions across the time dimension. Specifically, the input video sequence is compressed in the time dimension from $(1+T)$ frames to $[1+T/4]$, where the temporal resolution is reduced by a factor of 4. This temporal compression allows the model to learn key temporal features while reducing the computational cost of processing long video sequences.

The 3D VAE encoding process can be formally expressed as:
\begin{equation}
q(z | V) = \mathcal{N}(z; \mu, \sigma^2)
\end{equation}
where $\mu$ and $\sigma$ are the mean and variance outputs from the VAE encoder, and $z$ represents the latent variables.

\subsection{Diffusion Transformer (DiT)}
\label{app:dit}

After obtaining the latent representations from the 3D VAE encoder, these are passed to the Diffusion Transformer (DiT) to model the temporal evolution of the video. The DiT is responsible for predicting future frames of the video through a denoising diffusion process.

The input to the DiT consists of a sequence of latents $\{z_0, z_1, \dots, z_T\}$, where each $z_t$ represents the concatenated latent representation of the video and heatmap at time step $t$. During inference, Gaussian noise is added to these latent representations to simulate the diffusion process. The model is trained to denoise the noisy latents at each timestep.

The model predicts the velocity between latent frames using flow matching~\cite{lipman2022flow}. This velocity prediction is crucial for generating realistic video frames. The objective is to minimize the difference between the predicted velocity and the ground truth, defined as:
\begin{equation}
    v_t = \frac{d z_t}{d t} = z_1 - z_0
\end{equation}
During inference, the model predicts the velocity at each timestep to generate temporally consistent video sequences.

\subsection{3D VAE Decoder}
\label{app:vae_decoder}

After the latents are processed by the DiT, they are passed to the 3D VAE decoder to reconstruct the video frames. The decoder takes the output latent representations $\{z_0, z_1, \dots, z_T\}$ and maps them back to the pixel space, producing the final RGB or heatmap sequences $\hat{V} \in \mathbb{R}^{(1+T) \times H \times W \times 3}$.

The decoding process mirrors the encoding process, with the addition of upsampling operations to reconstruct high-resolution frames from the low-dimensional latent space.

\subsection{Efficient Inference with Cache Mechanism}
\label{sec:cache_infer}

To facilitate efficient inference for long video sequences, Wan incorporates a feature cache mechanism that stores previously computed features for reuse during the inference process. This approach reduces both memory and computational requirements, enabling the model to handle arbitrarily long video sequences.

For further details on the diffusion process, we refer the reader to~\cite{wan2025}.

\section{Projection and Back Projection}
\label{app.proj_back_proj}

We adopt the projection and back projection techniques outlined in BridgeVLA~\cite{li2025bridgevla} and RVT-2~\cite{goyal2024rvt}, with the following brief overview of the process:

\subsection{Why Virtual-View Rendering?}
\label{app:why_virtual_views}

\paragraph{Configurable and task-relevant viewpoints.}
Virtual-view rendering decouples the model input from the placement of the physical cameras. Raw RGB-D images are restricted to the viewpoints determined by sensor installation, whereas the fused point cloud can be rendered from virtual viewpoints that more clearly expose the workspace and task-relevant interaction regions. The viewing directions, rendering bounds, and spatial resolution can therefore be configured for a given workspace and remain fixed during training and evaluation.

\begin{figure*}[t]
  \centering
  \includegraphics[width=0.82\textwidth]{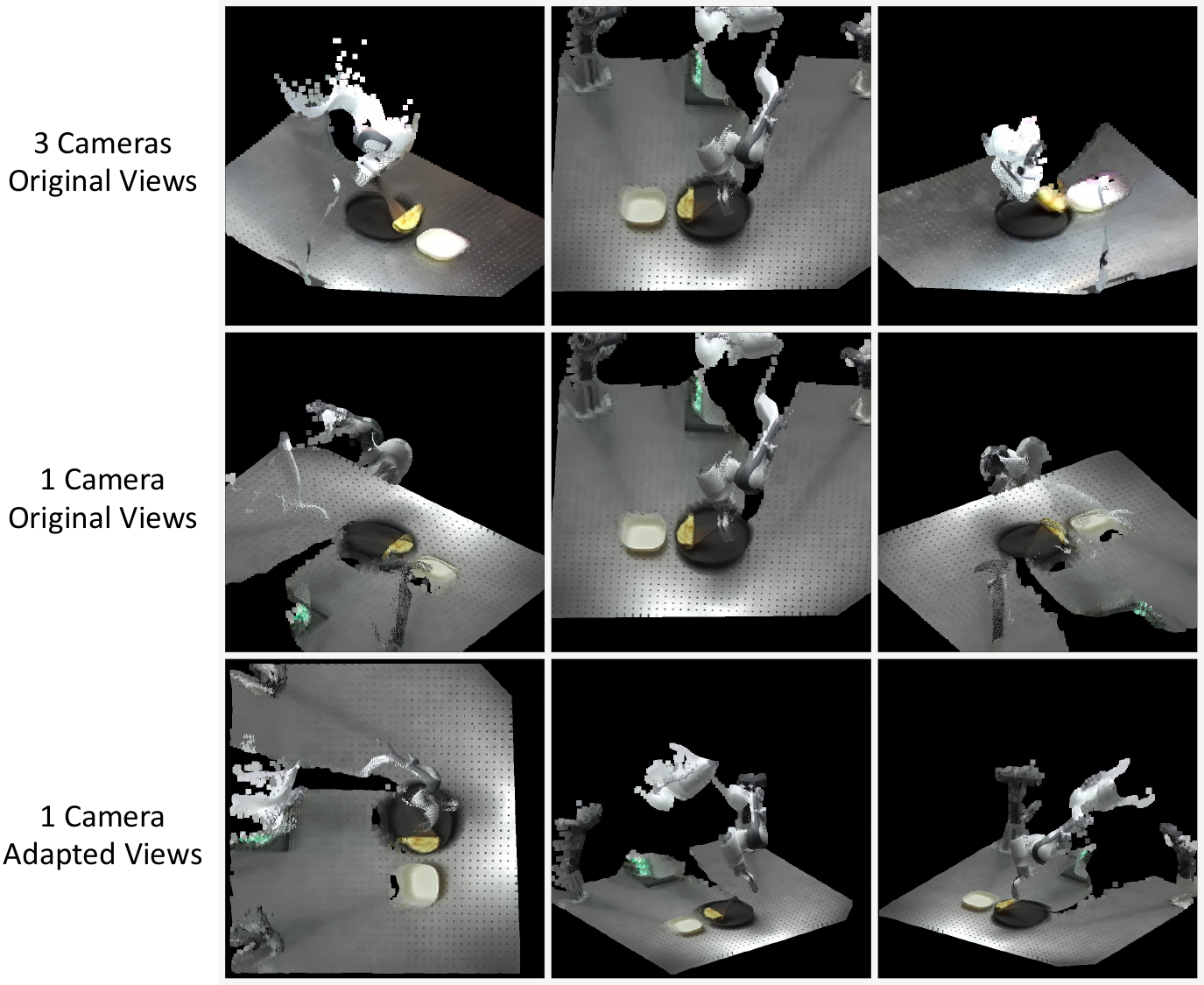}
  \caption{\textbf{Projected observations under different physical-camera and virtual-view configurations.} The three rows show, from top to bottom: three physical cameras with the original virtual projection views, one physical camera with the original views, and one physical camera with adapted views. Although using one camera reduces point-cloud coverage, adapting the virtual views better exposes the workspace observable from the remaining camera.}
  \label{fig:camera_view_visualization}
\end{figure*}

\paragraph{Multi-view representations with fewer physical cameras.}
Virtual rendering also preserves a multi-view input representation when fewer physical cameras are available. A single RGB-D camera produces a less complete point cloud, and rendering cannot recover surfaces that were never observed. Nevertheless, the observed geometry can still be presented from multiple virtual directions, allowing the model to reason over complementary projections of the available 3D information. As shown in Table~\ref{tab:camera_number_robustness} and Fig.~\ref{fig:camera_view_visualization}, adapting the virtual views to the workspace observable from one camera yields performance comparable to the original three-camera setting.

\paragraph{Geometrically consistent augmentation.}
Point-cloud rendering naturally supports consistent 3D data augmentation. During training, we apply the same SE(3) transformation to the point cloud and end-effector trajectory before rendering the transformed observations and heatmaps from the same virtual viewpoints. In contrast, independently applying 2D transformations to raw camera images does not generally preserve the underlying 3D geometry, cross-view consistency, or alignment with the transformed action trajectory.

\subsection{Projection Process}
The projection procedure involves three primary steps for mapping a point cloud of $N$ points to an RGB and depth image of size $(h,w)$:

\paragraph{Projection}
To project a 3D point cloud onto a 2D image, we first compute the depth $d_n$ and the pixel coordinates $(x_n, y_n)$ for each point indexed by $n \in \{0, 1, \dots, N\}$, where $f_n$ is the RGB value. These values are derived using the camera's intrinsic and extrinsic parameters. The 2D pixel index $i_n$ is then calculated as $i_n = x_n \cdot w + y_n$, where $w$ is the image width. This projection process can be efficiently accelerated by leveraging GPU-optimized matrix multiplications.

\paragraph{Z-ordering}
For each pixel $j$ in the image, we identify the point with the smallest depth $d_n$ among all the points projecting onto that pixel $\{n \mid i_n = j\}$. The RGB value $f_n$ of this point is then assigned to pixel $j$ in the RGB image, and the corresponding depth $d_n$ is stored in the depth image.

To enhance the performance of Z-ordering, we pack the depth and index of each point into a 64-bit integer. The higher 32 bits store the depth, and the lower 32 bits store the point index. This packing allows us to use two CUDA kernels for efficient computation. In the first kernel, we process each point in parallel and attempt to store its packed depth-index value at the corresponding pixel $j$ in a depth-index image using the \textit{atomicMin} operation. This ensures that only the point with the smallest depth at each pixel is retained. In the second kernel, the depth-index is unpacked, and the depth and feature images are reconstructed by referencing the corresponding point features. This technique, initially introduced in~\citep{schutz:cgf2021} for rendering colored point clouds by packing 32-bit color values, is extended here to support images with an arbitrary number of channels.

\paragraph{Screen-Space Splatting}
While the projection and Z-ordering steps are sufficient for basic image rendering, points are treated as infinitesimal light sources, which can result in visual noise when the resolution of the point cloud in screen space is lower than the image resolution. To address this issue, we apply 3D splatting, where each point is modeled as a disc of radius $r$ facing the camera. This approach is applied in screen space after projection and Z-ordering, reducing the computational burden of these steps. For each pixel $j$, the algorithm searches within a local neighborhood to find the nearest pixel $k$ with a smaller depth. If pixel $k$ satisfies $d_k < d_j$ and is closer than $r \cdot \text{focal\_length} / d_k$, the feature and depth of pixel $j$ are replaced by those of pixel $k$.

\subsection{Back Projection Process}
The back-projection procedure aims to identify the most accurate 3D position within the workspace, based on the three heatmaps generated by our virtual cameras. This process discretizes the workspace into a set of 3D points, computing the corresponding locations for each point on the three heatmaps. The probabilities for each point across the three heatmaps are averaged, and the point with the highest average probability is selected as the target 3D point.

For a more detailed explanation of these techniques, we refer the reader to the original works~\cite{li2025bridgevla,goyal2024rvt}.

\section{Training \& Inference Details}
\label{app:details on train and infer}

In this section, we detail the training and inference procedures for \method{}. The corresponding hyperparameters used during both training and inference are listed in Table~\ref{tab:hyperparameters}.

\input{tables/hyperparameters}

For the Meta-World benchmark, we train \method{} using a total of 32 NVIDIA H20 GPUs for 21.8k steps. The training of the multi-view video diffusion transformer and the Rotation and Gripper Predictor takes approximately 42 hours and 36 hours, respectively. For the real-world benchmark, we train the same models using 32 NVIDIA H200 GPUs for 8.6k steps, which takes around 11 hours and 10 hours, respectively.

For real-world deployment, the inference process requires more than 30 GB of GPU memory. Thus, we perform inference on a server with an NVIDIA A100 GPU, utilizing FastAPI for communication between the server and the robot client. To assess inference time, we conduct 30 trials, and the average time taken from point cloud input to the output of an action chunk (of length 24) is 4.6 seconds.

\section{Deployment Robustness Analysis}
\label{app:camera_calibration_robustness}
We further examine whether the default use of multiple fixed and calibrated physical cameras is essential to \method{}. Following the real-world evaluation protocol in Sec.~\ref{exp:real_world}, we separately study robustness to reduced camera coverage and calibration errors across the same seven manipulation tasks.

\subsection{Robustness to Camera Number}
In the default setting, observations from three physical cameras are fused into a colored point cloud. For the single-camera settings, we retain only one of these cameras, substantially reducing point-cloud coverage, and compare the original virtual projection views with adapted views selected to better cover the workspace observable from the remaining camera. All three settings are evaluated on the same seven real-world tasks with 10 trials per task.

\input{tables/camera_number_robustness}

\subsection{Robustness to Calibration Errors}
We randomly perturb the camera extrinsics by $20\,\mathrm{mm}$ in translation and $2^\circ$ in rotation. Under Setting~1, training uses the original calibration while evaluation uses the perturbed calibration, introducing a train--test mismatch. Under Setting~2, training and evaluation use the same perturbed calibration, representing a consistent calibration bias. We compare both settings with the original calibration on the same seven real-world tasks, using 10 trials per task.

\input{tables/calibration_robustness}

\section{Robustness Analysis for Key Parameters}
\label{app:robustness_analysis}
\input{tables/hyperparameter_robustness}

In this section, we evaluate the robustness of our model with respect to key hyperparameters, specifically the RGB loss weight $\lambda$ and Heatmap standard deviation $\sigma$. We vary these hyperparameters individually, retrain the model, and re-evaluate it on the Meta-World benchmark. To ensure a fair comparison, we keep all other parameters fixed for each experiment. The results are summarized in Table~\ref{tab:robustness_analysis}. As shown, when the $\lambda$ parameter is varied by 80\%, the success rate changes by only 2.9\%; when the $\sigma$ parameter is varied by 133\%, the success rate changes by just 2.2\%. These findings highlight that our model is highly robust to variations in these hyperparameters.

\section{Meta-World Simulation Baselines}
\label{app:sim_baselines}

We evaluate our method against a diverse set of imitation-learning and video-prediction baselines, commonly benchmarked on the Meta-World dataset. The baseline results presented in our tables follow the numbers from prior works~\cite{ko2023learning,Chen_2025_ICCV}. Below, we summarize the core characteristics and implementations of each baseline.

\paragraph{BC-Scratch and BC-R3M}  
BC-Scratch~\cite{nair2022r3m} is a standard multi-task behavioral cloning baseline, trained end-to-end using expert demonstrations. It employs a dual-stream encoder, processing RGB observations with a ResNet-18 and task descriptions with a CLIP text encoder. The encoded features are concatenated and passed through a multi-layer perceptron to directly predict actions. BC-R3M~\cite{nair2022r3m} follows the same architecture and training protocol but initializes the visual encoder with R3M-pretrained weights, which improves visual representations and generalization.

\paragraph{Diffusion Policy}  
Diffusion Policy~\cite{chi2025diffusion} frames action generation as a diffusion process, where continuous action sequences are sampled through iterative denoising, conditioned on a fixed window of past observations. Trained via imitation learning, this approach offers a flexible and expressive model of multimodal action distributions. However, it does not explicitly model future visual observations nor utilize 3D-related information.

\paragraph{UniPi}  
UniPi~\cite{du2023learning} is a two-stage video-prediction policy learning framework. It first learns a vision-language–conditioned video prediction model to forecast future observations, and then trains an inverse dynamics model using behavioral cloning to map predicted frame transitions to actions. This decomposition allows UniPi to leverage learned visual dynamics, though action generation remains indirect and dependent on the inverse dynamics model's quality.

\paragraph{AVDC}  
AVDC~\cite{ko2023learning} adopts an object-centric, flow-based approach for visuomotor control. It predicts future video frames and extracts object motion trajectories using off-the-shelf point tracking methods. These trajectories are then used to generate robot actions. AVDC leverages a flow-based formulation that decouples perception and control, which helps it achieve robust performance across a range of environments.

\paragraph{Track2Act}  
Track2Act~\cite{bharadhwaj2024track2act} is an object-centric manipulation framework that decomposes action generation into object motion prediction and control. Given visual observations and a goal specification, it first predicts future object trajectories by tracking keypoints in videos. These trajectories then condition a behavior cloning policy to output robot actions. To mitigate execution errors and modeling inaccuracies, Track2Act incorporates a residual action correction module that refines predicted actions during execution.

\paragraph{DreamZero.}
DreamZero~\cite{ye2026worldactionmodelszeroshot} is a 14B video-action model (VAM) built on a pretrained video diffusion backbone. It jointly predicts future video frames and robot actions in an autoregressive manner using flow matching with teacher forcing. Through a series of system- and model-level optimizations, including DreamZero-Flash, which decouples the denoising schedules for video and action prediction, DreamZero achieves a 38$\times$ inference speedup, enabling real-time closed-loop control at 7\,Hz. For a fair comparison with our method, we replace its original Wan2.1-I2V-14B backbone with Wan2.2-TI2V-5B to better align the two methods in model capacity. All other settings follow the official recommendations.

\section{Real-World Baselines}
\label{app:real_world_baselines}

In real-world experiments, we compare \method{} with representative baselines spanning 3D-based visuomotor policies, video-prediction methods, and vision-language-action (VLA) models. Below, we describe the core modeling choices of each baseline and how they are instantiated in our setting.

\paragraph{DP3}
DP3~\cite{ze20243d} is a 3D-based visuomotor policy that directly maps geometric observations to continuous action sequences. It represents the environment using sparse point clouds reconstructed from depth observations, which are encoded with a lightweight MLP-based point encoder. Actions are generated using a diffusion-based policy head conditioned on the encoded 3D features and robot proprioception.

In contrast to \method{}, DP3 does not leverage the pretrained knowledge of foundation models, nor does it explicitly predict future visual observations. For fairness in our experiments, we use the official DP3 implementation with point clouds constructed from multi-view depth observations.

\paragraph{$\boldsymbol{\pi_{0.5}}$}
$\pi_{0.5}$~\cite{intelligence2025pi_}   is a large-scale vision-language-action (VLA) model that unifies perception, language understanding, and action generation within a single autoregressive transformer. It formulates imitation learning as next-token prediction, jointly processing visual observations, language instructions, and robot actions using a pretrained vision-language backbone. One key feature of $\pi_{0.5}$ is its hybrid action representation. During training, actions are discretized into tokens for scalable pretraining, while at inference time, an action expert generates continuous action chunks via flow matching. This design enables efficient non-autoregressive action generation while retaining the language grounding and reasoning capabilities of the underlying model. However, $\pi_{0.5}$ does not explicitly model future visual observations or incorporate 3D geometric representations. 
In our experiments, we use the official $\pi_{0.5}$ implementation, inputting 3 RGB observations for fair comparison.

\paragraph{UVA}
UVA~\cite{li2025unified}   is a video-prediction policy learning framework that jointly models future visual observations and robot actions. It learns a unified latent representation by predicting future video frames and corresponding action chunks over a fixed horizon, encouraging the model to capture scene dynamics. While video and action prediction are trained jointly, policy inference decodes actions directly from the learned latent representation for efficiency. Despite modeling future observations, UVA operates on single-view video inputs and does not leverage multi-view prediction or cross-view geometric consistency. In our experiments, we follow the default settings of UVA, using 4-step history observations and single-view video prediction.

\paragraph{BridgeVLA}
BridgeVLA~\cite{li2025bridgevla} is a 3D-aware vision-language-action model that predicts robot actions via heatmap-based keyframe estimation. It represents the scene using point clouds reconstructed from RGB-D observations, which are projected into multiple orthographic views. A vision-language backbone predicts 2D heatmaps for each view, which are back-projected to infer the next end-effector translation. Remaining action components are predicted using an MLP. Action execution is carried out by an external motion planner that moves the robot to the predicted keyframe pose.

BridgeVLA is trained using large-scale 2D heatmap pretraining, followed by 3D action fine-tuning, aligning vision-language pretraining with downstream manipulation tasks. However, it predicts only a single next keyframe at each step and does not explicitly model continuous action execution or future visual evolution. We evaluate BridgeVLA using its official implementation and merged point clouds from multiple depth cameras.

Overall, these baselines cover a broad spectrum of design choices, ranging from direct 3D-to-action mappings, to video prediction, to vision--language–guided key-pose estimation, enabling a comprehensive comparison with our multi-view video diffusion policy.

\section{Detailed Real-World Analysis}
\label{app:real_world_detail}

We provide the per-baseline qualitative failure analyses referenced in Sec.~\ref{exp:real_world}, as well as the primary failure mode of \method{}.

\paragraph{DP3.}
DP3 severely overfits to the training set. On the \textit{Put Lion} task, despite its training loss decreasing from $10^{-1}$ to $10^{-8}$, the policy fails to approach the target object, instead directly moving toward target regions (the shelf). This failure is attributed to DP3's simple MLP-based point cloud encoder, which struggles with sensor noise and real-world variability.

\paragraph{$\boldsymbol{\pi_{0.5}}$.}
For $\pi_{0.5}$, spatial generalization is limited. On \textit{Put Lion}, it can only approach the target within the training distribution; otherwise, it fails. On \textit{Push T}, it completely fails, showing no success tendency. On \textit{Scoop Tortilla}, it scoops the tortilla reliably but fails to place it on the plate, likely due to plate placement variations in the training data.

\paragraph{UVA.}
UVA exhibits reasonable action tendencies across all tasks, but its execution accuracy is insufficient. On \textit{Put Lion}, it frequently closes the gripper prematurely. On \textit{Push T}, it tends to over-push the object, entering out-of-distribution states. On \textit{Scoop Tortilla}, the dominant failure mode is inaccurate pouring, where the tortilla often falls outside the plate.

\paragraph{BridgeVLA.}
BridgeVLA is the strongest competing baseline. Similar to our approach, it projects 3D point clouds into multi-view images and predicts heatmaps before decoding actions. However, it is based on a vision--language model (PaliGemma~\citep{beyer2024paligemma}) pretrained on static image--text pairs, which cannot predict continuous video or heatmap sequences. As a result, it predicts only a single key pose, requiring an external motion planner for actions between poses. This works for \textit{Put Lion}, but fails for \textit{Push T}, where defining key poses is difficult. For contact-rich tasks like \textit{Scoop Tortilla}, intermediate actions are crucial for success, and predicting only a final key pose is insufficient. We observe that BridgeVLA often fails in \textit{Scoop Tortilla}, even with a reasonable key pose.

\paragraph{\method{} failure mode.}
The primary failure mode of \method{} stems from inaccurate action decoding. For example, on \textit{Scoop Tortilla}, the policy occasionally scrapes across the top surface instead of lifting the tortilla from underneath. We attribute this limitation to the finite resolution of the heatmaps ($256 \times 256$), where each pixel corresponds to approximately $4\,\mathrm{mm}$. Increasing the heatmap resolution may further improve performance.

\section{Visualization of Video Prediction}
\label{app:vis_video}
\begin{figure*}[t]
  \centering
  \includegraphics[width=0.7\textwidth]{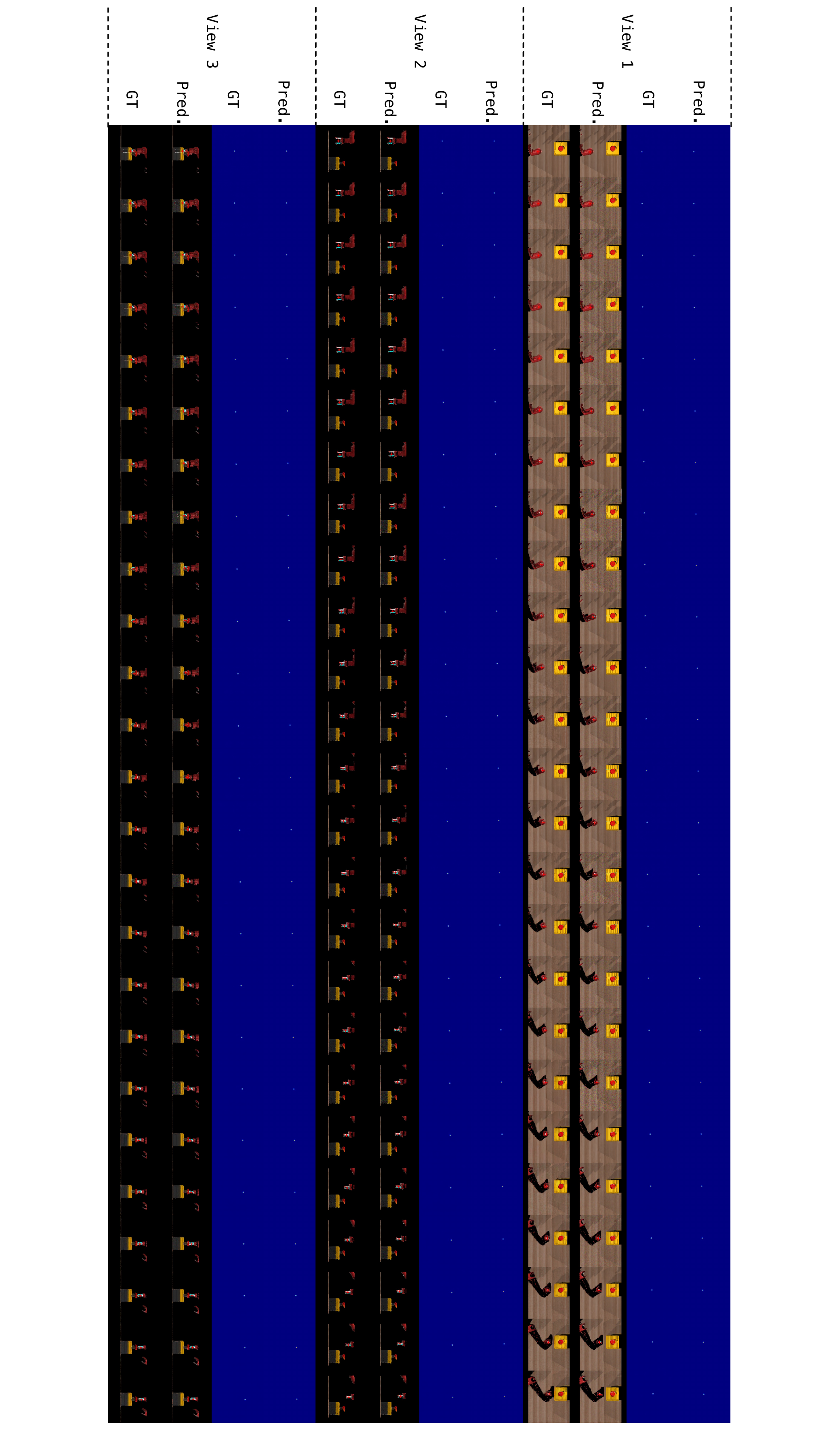}
  \vspace{-0.2cm}
\caption{
\textbf{Visualization of the predicted RGB sequences and heatmap sequences for the \textit{Button-Press-Top} task in Meta-World.} For each view, the first and third rows show predictions from \method{}, while the second and fourth rows show the corresponding ground truth. The peak locations of both predicted and ground-truth heatmaps are overlaid on the predicted and ground-truth RGB images, respectively. The results show that (1) the predicted RGB sequences are visually realistic, and (2) the heatmap peaks closely follow the end effector in the RGB images, indicating strong consistency between the predicted heatmaps and RGB sequences.
}
  \vspace{-0.3cm}
  \label{fig:button_press_top}
\end{figure*}

\begin{figure*}[t]
  \centering
  \includegraphics[width=0.7\textwidth]{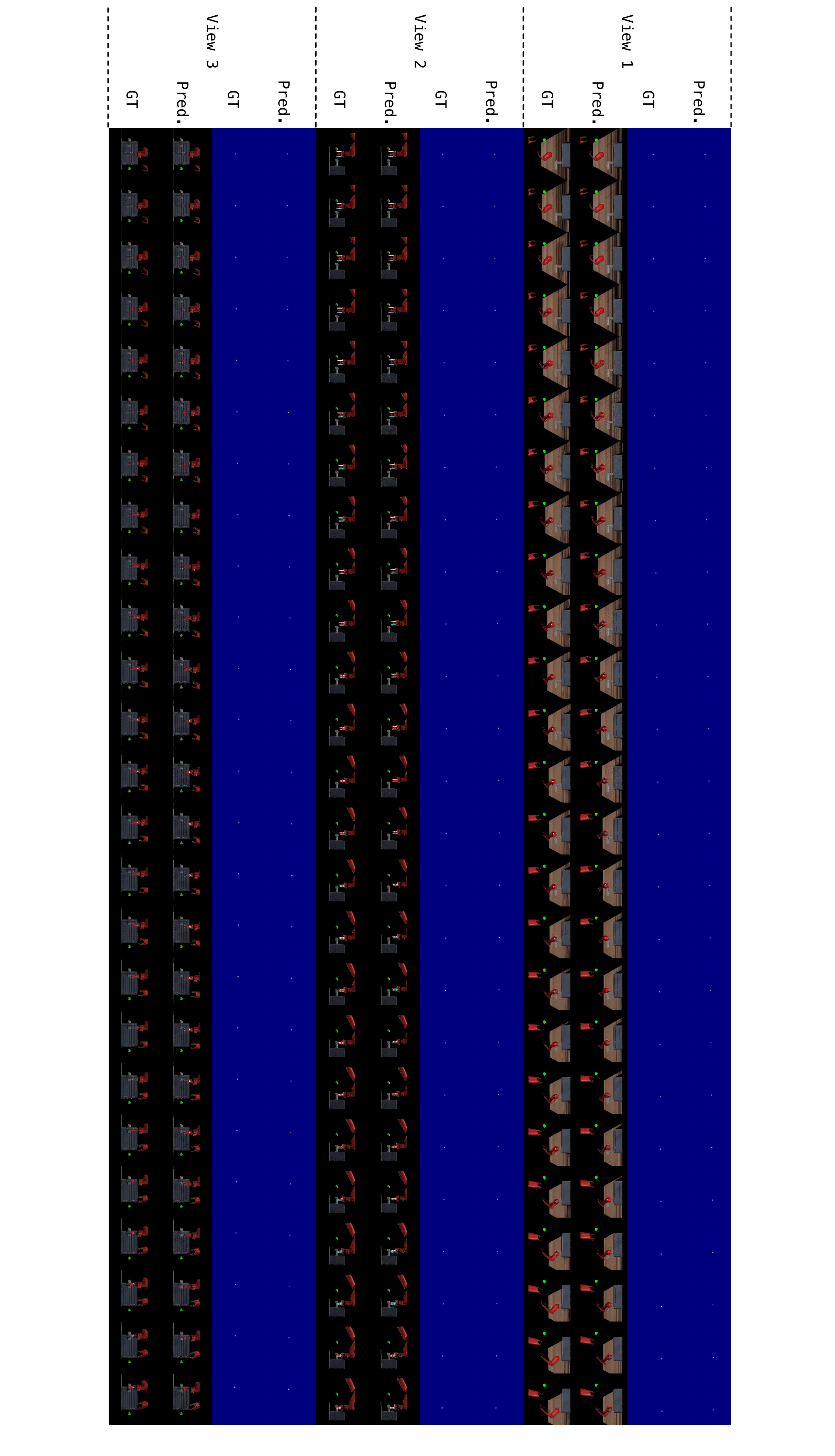}
  \vspace{-0.2cm}
\caption{
\textbf{Visualization of the predicted RGB sequences and heatmap sequences for the \textit{Door-Open} task in Meta-World.} For each view, the first and third rows show predictions from \method{}, while the second and fourth rows show the corresponding ground truth. The peak locations of both predicted and ground-truth heatmaps are overlaid on the predicted and ground-truth RGB images, respectively. The results show that (1) the predicted RGB sequences are visually realistic, and (2) the heatmap peaks closely follow the end effector in the RGB images, indicating strong consistency between the predicted heatmaps and RGB sequences.
}
  \vspace{-0.3cm}
  \label{fig:door_open}
\end{figure*}

\begin{figure*}[t]
  \centering
  \includegraphics[width=0.7\textwidth]{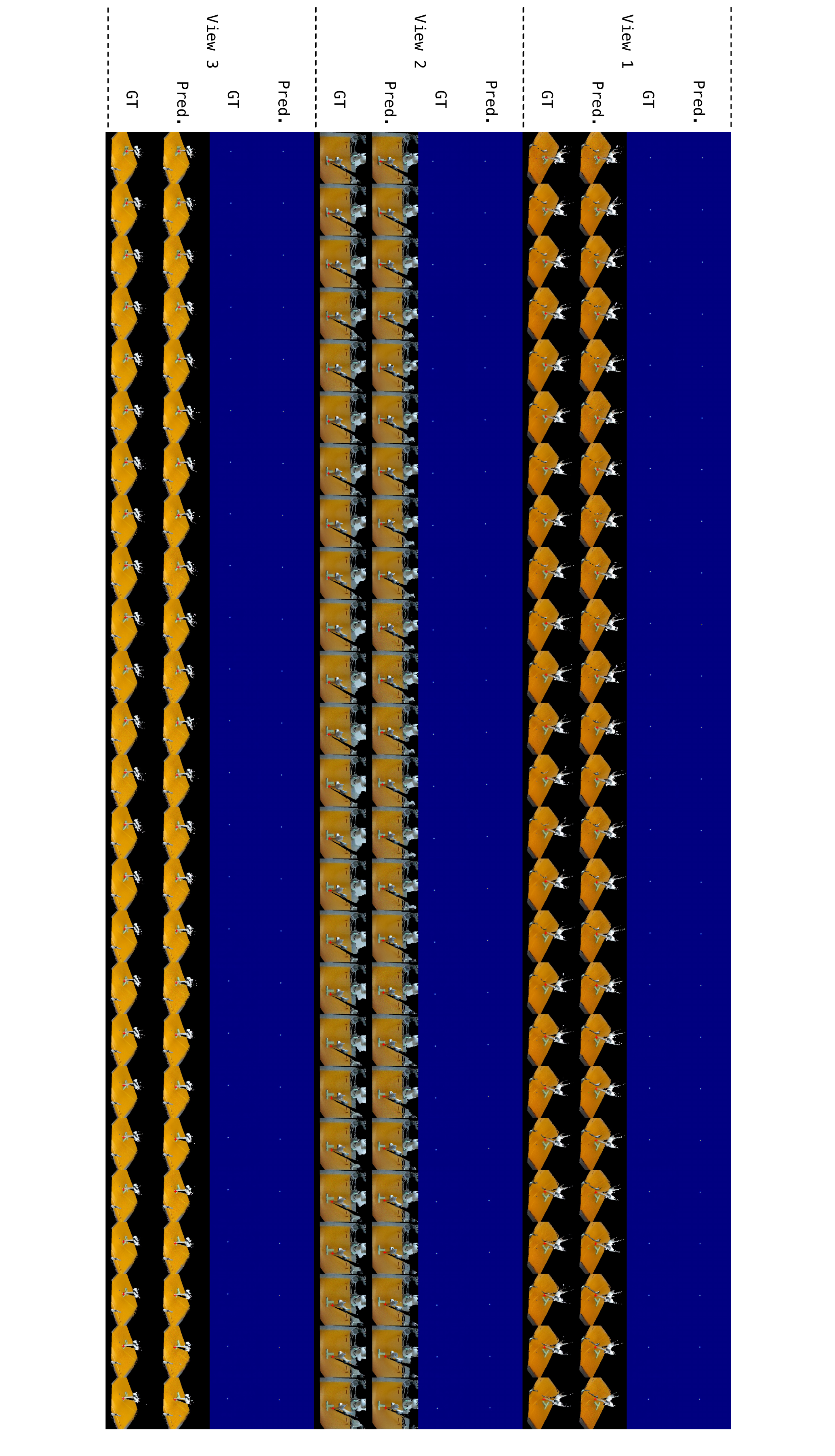}
  \vspace{-0.2cm}
\caption{
\textbf{Visualization of the predicted RGB sequences and heatmap sequences for the \textit{Push-T} task.} For each view, the first and third rows show predictions from \method{}, while the second and fourth rows show the corresponding ground truth. The peak locations of both predicted and ground-truth heatmaps are overlaid on the predicted and ground-truth RGB images, respectively. The results show that (1) the predicted RGB sequences are visually realistic, and (2) the heatmap peaks closely follow the end effector in the RGB images, indicating strong consistency between the predicted heatmaps and RGB sequences.
}
  \vspace{-0.3cm}
  \label{fig:push_T_2}
\end{figure*}

\begin{figure*}[t]
  \centering
  \includegraphics[width=0.7\textwidth]{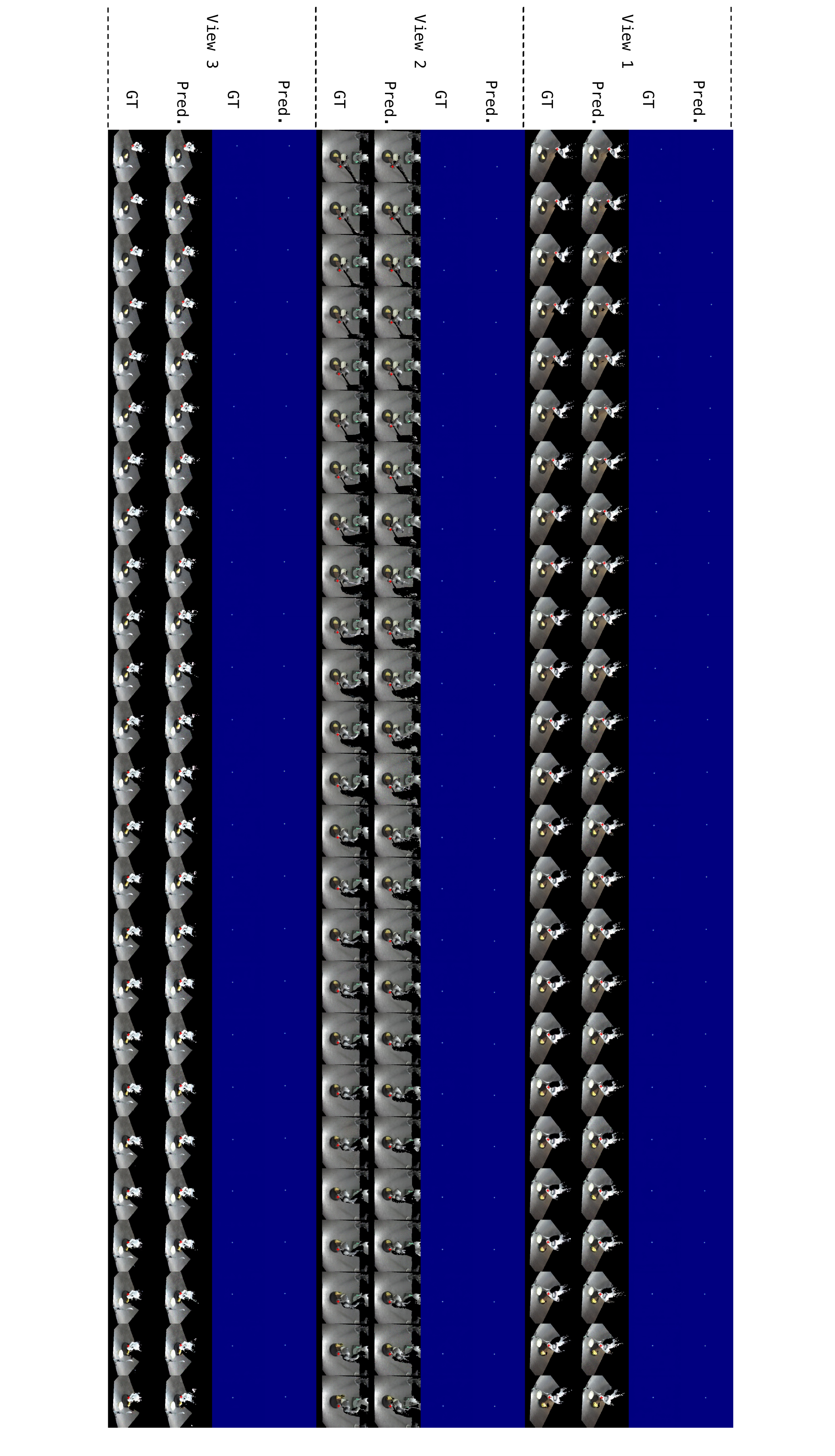}
  \vspace{-0.2cm}
\caption{
\textbf{Visualization of the predicted RGB sequences and heatmap sequences for the \textit{Scoop Tortilla} task.} For each view, the first and third rows show predictions from \method{}, while the second and fourth rows show the corresponding ground truth. The peak locations of both predicted and ground-truth heatmaps are overlaid on the predicted and ground-truth RGB images, respectively. The results show that (1) the predicted RGB sequences are visually realistic, and (2) the heatmap peaks closely follow the end effector in the RGB images, indicating strong consistency between the predicted heatmaps and RGB sequences.
}
  \vspace{-0.3cm}
  \label{fig:cook_2}
\end{figure*}

\begin{figure*}[t]
  \centering
  \includegraphics[width=0.7\textwidth]{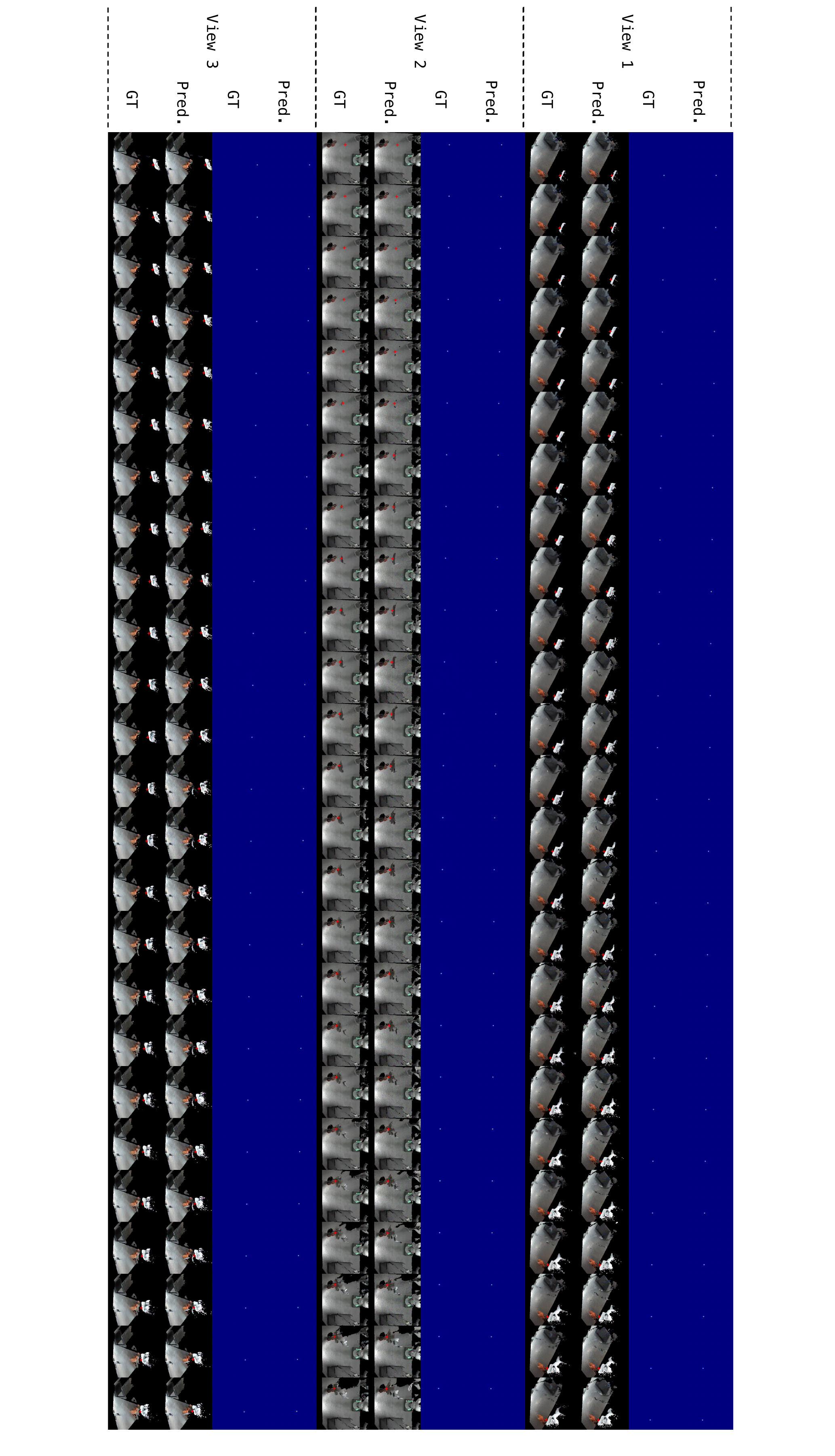}
  \vspace{-0.2cm}
\caption{
\textbf{Visualization of the predicted RGB sequences and heatmap sequences for the \textit{Put Lion} task.} For each view, the first and third rows show predictions from \method{}, while the second and fourth rows show the corresponding ground truth. The peak locations of both predicted and ground-truth heatmaps are overlaid on the predicted and ground-truth RGB images, respectively. The results show that (1) the predicted RGB sequences are visually realistic, and (2) the heatmap peaks closely follow the end effector in the RGB images, indicating strong consistency between the predicted heatmaps and RGB sequences.
}
  \vspace{-0.3cm}
  \label{fig:put_lion_2}
\end{figure*}

\begin{figure*}[t]
  \centering
  \begin{subfigure}[t]{1.0\textwidth}
    \centering
    \includegraphics[width=\textwidth]{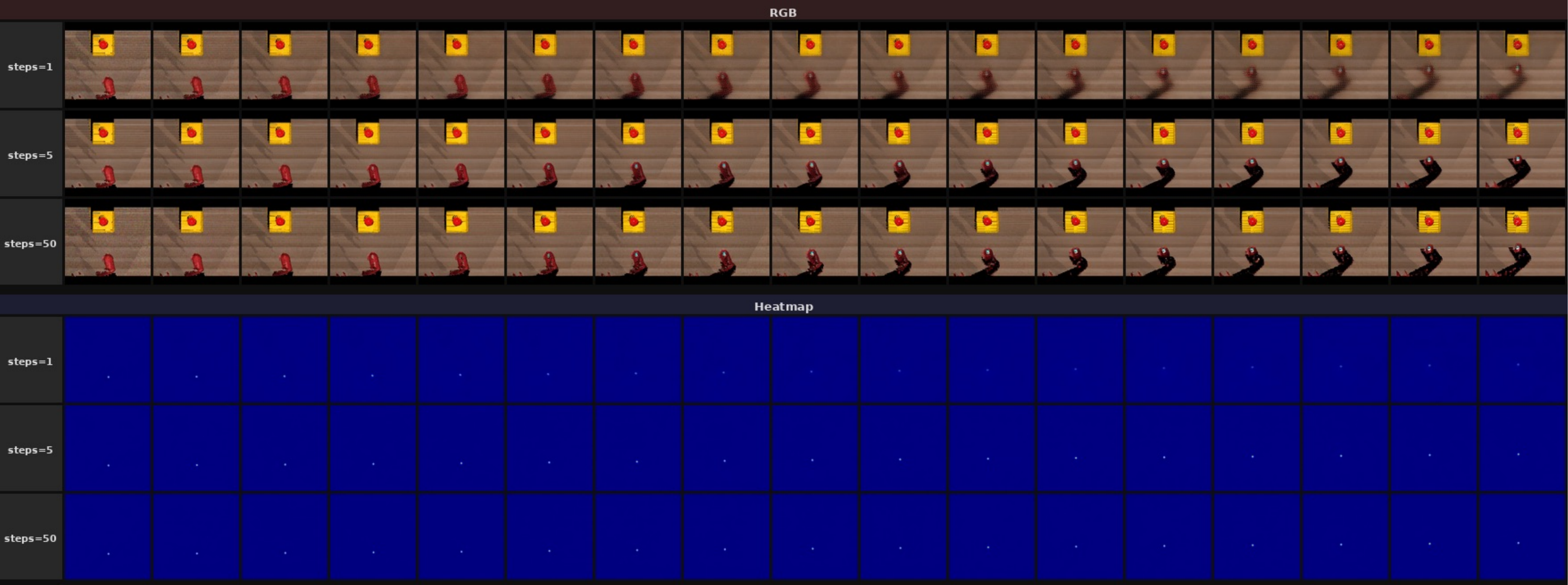}
  \end{subfigure}

  \vspace{0.2cm}

  \begin{subfigure}[t]{1.0\textwidth}
    \centering
    \includegraphics[width=\textwidth]{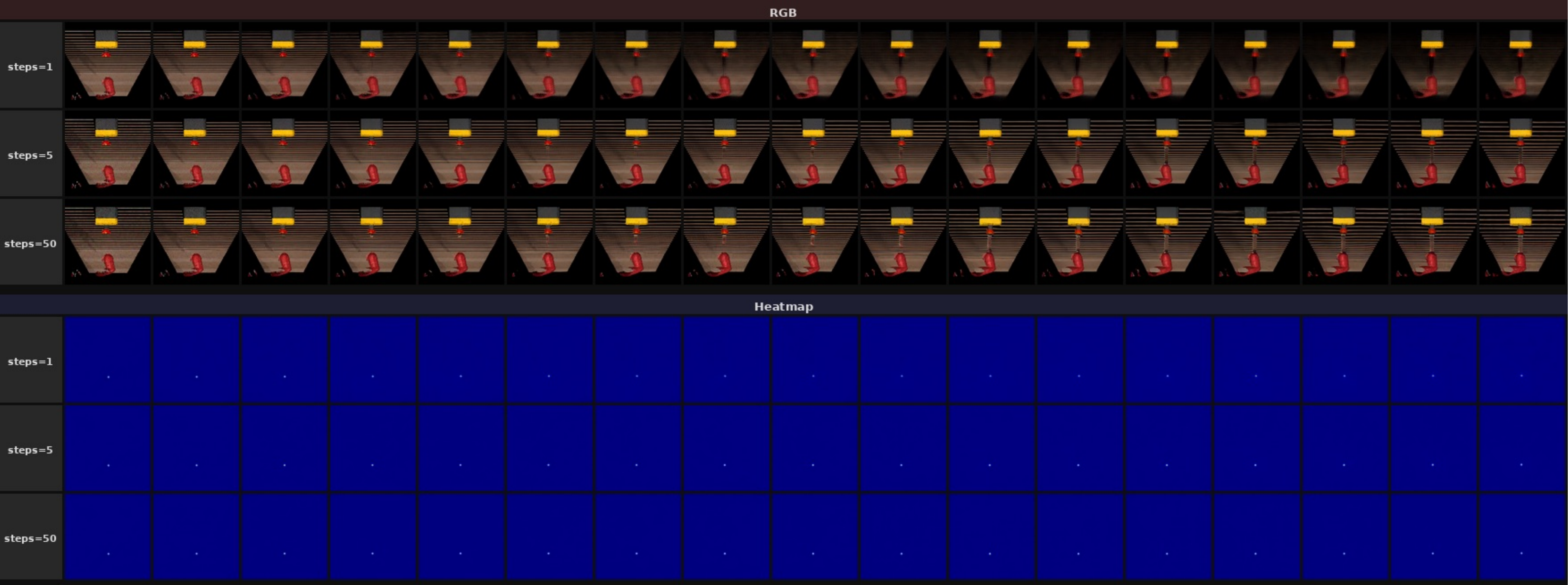}
  \end{subfigure}

  \vspace{0.2cm}

  \begin{subfigure}[t]{1.0\textwidth}
    \centering
    \includegraphics[width=\textwidth]{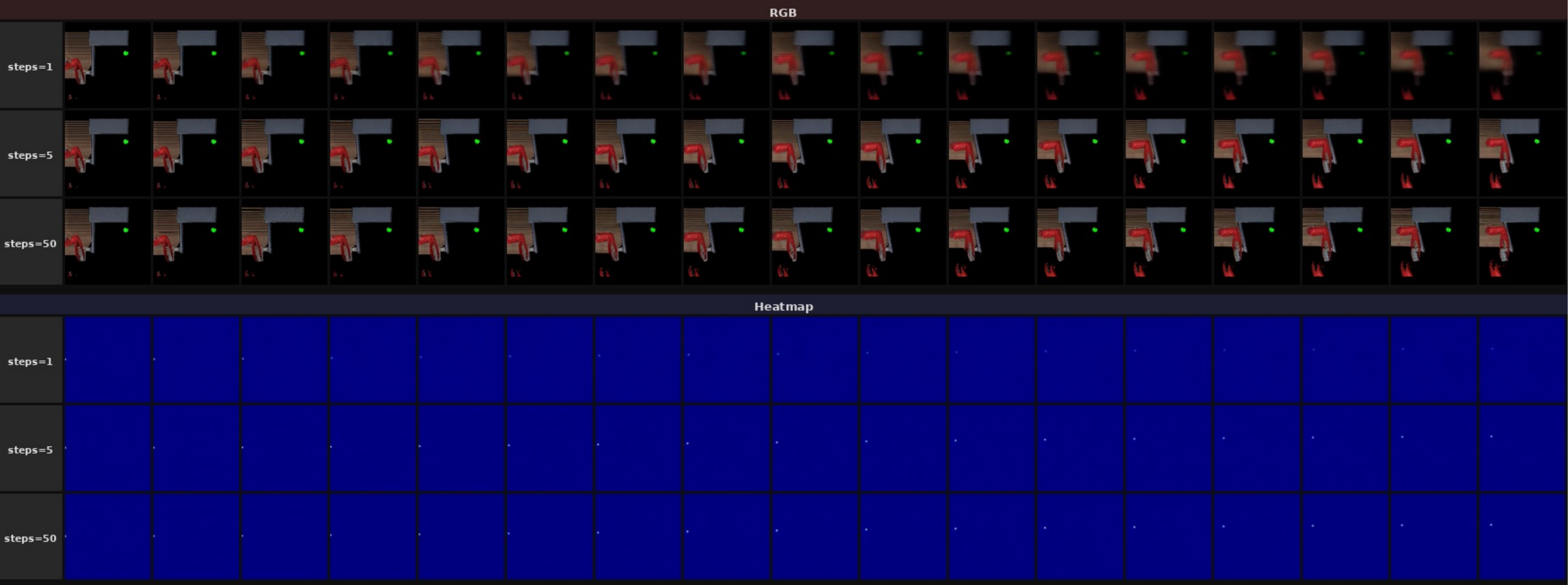}
  \end{subfigure}

\caption{\textbf{Visualization of video predictions under different denoising steps (Part I).} Predicted RGB videos and heatmap videos under different denoising step settings. Lower denoising steps lead to visibly lower RGB video quality, while the predicted heatmaps remain relatively stable.}
  \label{fig:compare_steps_partI}
\end{figure*}

\begin{figure*}[t]
  \centering
  \begin{subfigure}[t]{1.0\textwidth}
    \centering
    \includegraphics[width=\textwidth]{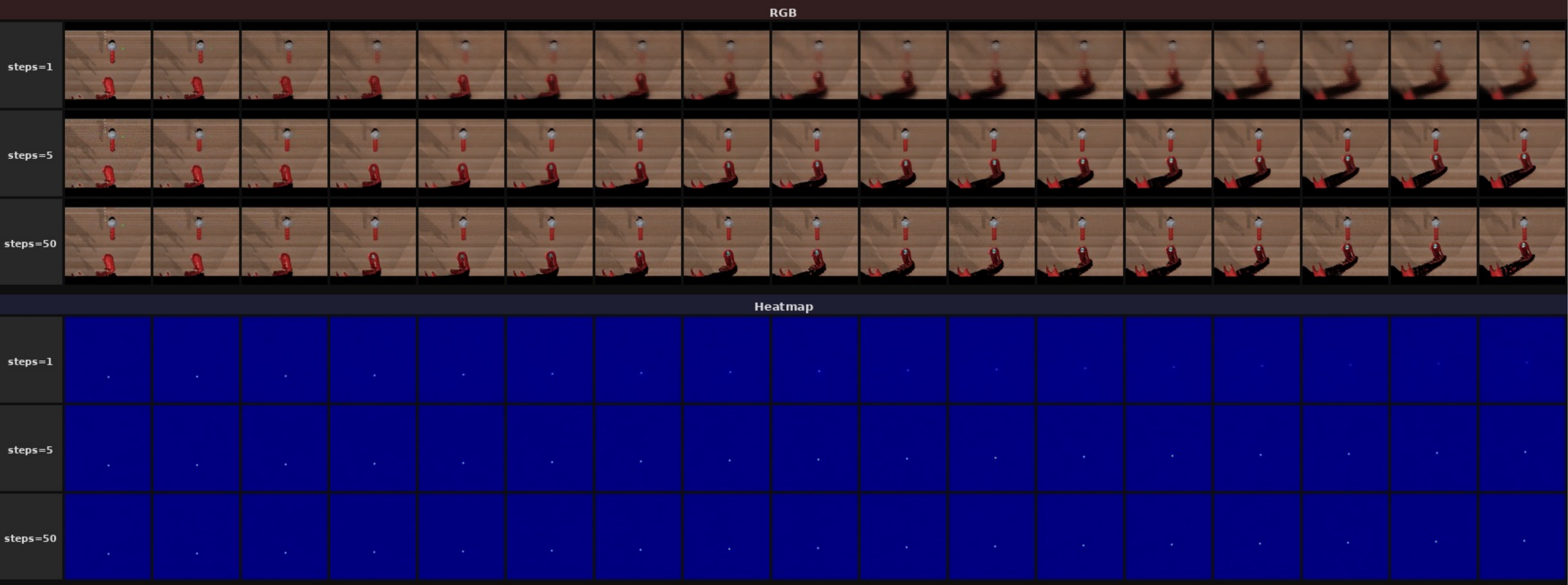}
  \end{subfigure}

  \vspace{0.2cm}

  \begin{subfigure}[t]{1.0\textwidth}
    \centering
    \includegraphics[width=\textwidth]{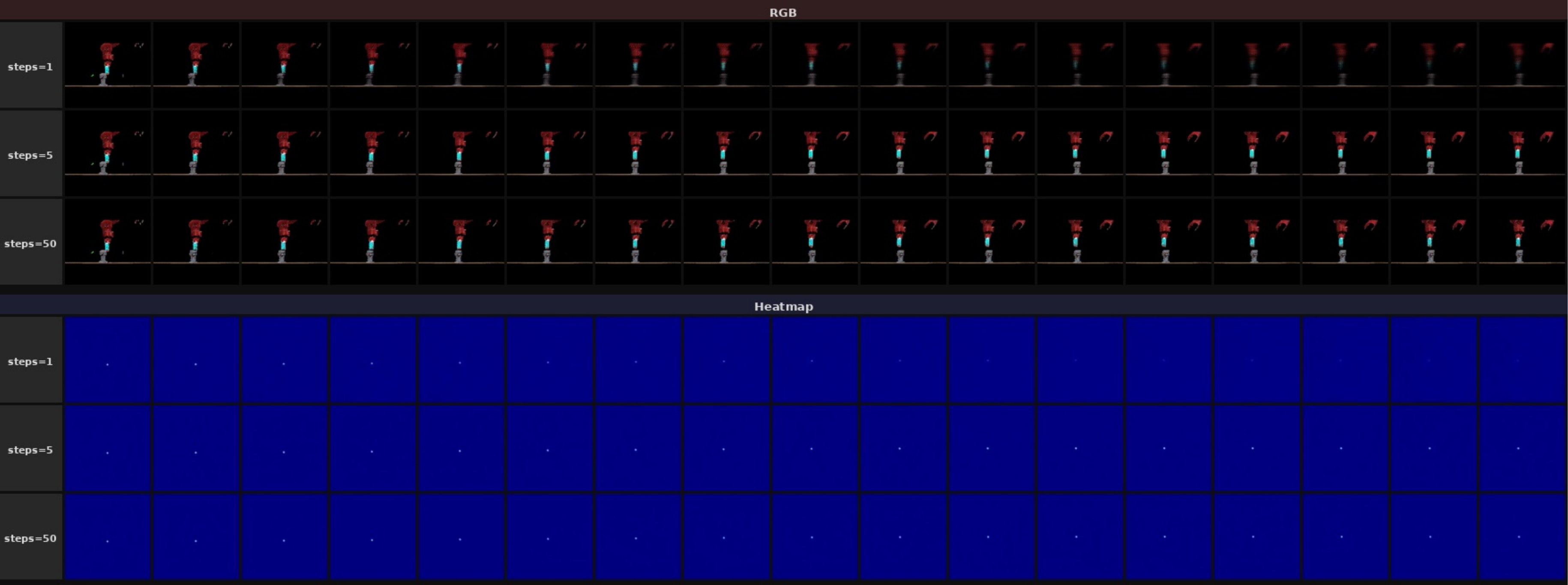}
  \end{subfigure}

  \vspace{0.2cm}

  \begin{subfigure}[t]{1.0\textwidth}
    \centering
    \includegraphics[width=\textwidth]{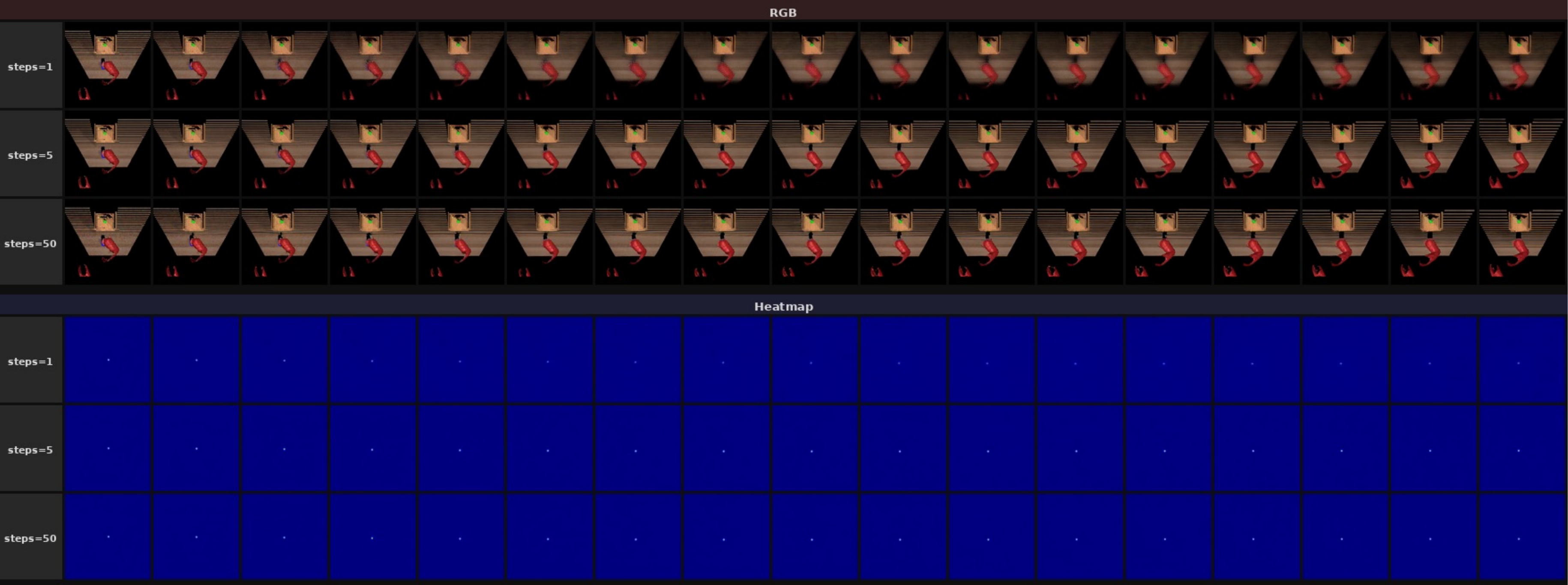}
  \end{subfigure}

\caption{\textbf{Visualization of video predictions under different denoising steps (Part II).} Predicted RGB videos and heatmap videos under different denoising step settings. Lower denoising steps lead to visibly lower RGB video quality, while the predicted heatmaps remain relatively stable.}
  \label{fig:compare_steps_partII}
\end{figure*}

We visualize the video prediction outputs from \method{} for both the Meta-World and real-world (Fig.~\ref{fig:button_press_top} to Fig.~\ref{fig:put_lion_2}) environments. As shown, the generated RGB videos are realistic and align well with the generated heatmaps. This consistency allows the predicted actions to be validated visually, providing an effective tool to avoid unsafe deployments.
We further compare the prediction results under different denoising step settings in Fig.~\ref{fig:compare_steps_partI} and Fig.~\ref{fig:compare_steps_partII}. When the denoising step is set to 1, the visual quality of the generated RGB videos deteriorates noticeably, while the peak locations of the predicted heatmap videos remain stable. Increasing the denoising step to 5 or 50 does not lead to a significant visual difference in the generated outputs.

\clearpage

\clearpage

%% file: tables/hyperparameters.tex
\begin{table*}[t]
\centering
\caption{Hyperparameters used for training and inference in \method{}.}
\label{tab:hyperparameters}

\small
\setlength{\tabcolsep}{8pt}
\renewcommand{\arraystretch}{1.15}

\begin{tabular}{lcc}
\toprule
\textbf{Hyperparameter} 
& \textbf{Multi-View Video Diffusion Transformer} 
& \textbf{Rotation \& Gripper Predictor} \\
\midrule
Optimizer & AdamW & AdamW \\
Learning rate & $1.0 \times 10^{-4}$ & $1.0 \times 10^{-4}$ \\
Epochs & 100 & 100 \\
Batch size & 64 & 64 \\
Weight decay & 0 & $1.0 \times 10^{-5}$ \\
Image resolution & $256 \times 256$ & $256 \times 256$ \\
Heatmap std. ($\sigma$) & 1.5 & -- \\
Predicted video length & 24 & -- \\
RGB loss weight  ($\lambda$) & 0.5 & -- \\
Heatmap loss weight & 0.5 & -- \\
LoRA modules & \texttt{q, k, v, o, ffn.0, ffn.2} & -- \\
LoRA rank & 32 & -- \\
Training timesteps & 1000 & -- \\
Inference steps & 5 & -- \\
Classifier-free guidance & 1 & -- \\
\bottomrule
\end{tabular}
\end{table*}

%% file: tables/camera_number_robustness.tex
\begin{table*}[t]
\centering
\caption{\textbf{Robustness to physical-camera coverage and virtual-view selection.} Success rates over 10 trials per task on three basic and four unseen tasks.}
\label{tab:camera_number_robustness}

\footnotesize
\setlength{\tabcolsep}{2.0pt}
\renewcommand{\arraystretch}{1.2}

\begin{tabularx}{\textwidth}{
>{\centering\arraybackslash}p{0.18\textwidth}
*{7}{>{\centering\arraybackslash}X}
>{\centering\arraybackslash}p{0.08\textwidth}
}
\toprule
\multirow{2}{*}[-0.5ex]{Setting}
& \multicolumn{3}{c}{Basic Tasks}
& \multicolumn{4}{c}{Unseen Tasks}
& \multirow{2}{*}[-0.5ex]{\makecell{Avg.\\Succ. (\%)}} \\
\cmidrule(lr){2-4}\cmidrule(lr){5-8}
& Put Lion & Push-T & Scoop Tort.
& Put-B & Put-H & Push-L & Scoop-C & \\
\midrule
Three Cameras
& \textbf{10/10} & 4/10 & 7/10
& \textbf{5/10} & 6/10 & \textbf{3/10} & \textbf{5/10}
& 57.1 \\
\midrule
\makecell{One Camera\\Original Views}
& 8/10 & 3/10 & 7/10
& \textbf{5/10} & 5/10 & \textbf{3/10} & 4/10
& 50.0 \\
\makecell{One Camera\\Adapted Views}
& \textbf{10/10} & \textbf{5/10} & \textbf{8/10}
& \textbf{5/10} & \textbf{7/10} & \textbf{3/10} & 4/10
& \textbf{60.0} \\
\bottomrule
\end{tabularx}

\vspace{-0.6em}
\end{table*}

%% file: tables/calibration_robustness.tex
\begin{table*}[t]
\centering
\caption{\textbf{Robustness to camera calibration errors.} Success rates over 10 trials per task under the original calibration and two perturbed-calibration settings.}
\label{tab:calibration_robustness}

\footnotesize
\setlength{\tabcolsep}{2.0pt}
\renewcommand{\arraystretch}{1.2}

\begin{tabularx}{\textwidth}{
>{\centering\arraybackslash}p{0.18\textwidth}
*{7}{>{\centering\arraybackslash}X}
>{\centering\arraybackslash}p{0.08\textwidth}
}
\toprule
\multirow{2}{*}[-0.5ex]{Setting}
& \multicolumn{3}{c}{Basic Tasks}
& \multicolumn{4}{c}{Unseen Tasks}
& \multirow{2}{*}[-0.5ex]{\makecell{Avg.\\Succ. (\%)}} \\
\cmidrule(lr){2-4}\cmidrule(lr){5-8}
& Put Lion & Push-T & Scoop Tort.
& Put-B & Put-H & Push-L & Scoop-C & \\
\midrule
\makecell{Original\\Calibration}
& \textbf{10/10} & \textbf{4/10} & 7/10
& 5/10 & \textbf{6/10} & 3/10 & \textbf{5/10}
& \textbf{57.1} \\
\midrule
\makecell{Perturbed Calibration\\Setting 1}
& 8/10 & 2/10 & 6/10
& \textbf{6/10} & 5/10 & \textbf{4/10} & \textbf{5/10}
& 51.4 \\
\makecell{Perturbed Calibration\\Setting 2}
& \textbf{10/10} & \textbf{4/10} & \textbf{8/10}
& 4/10 & 5/10 & 3/10 & 4/10
& 54.3 \\
\bottomrule
\end{tabularx}

\vspace{-0.6em}
\end{table*}

%% file: tables/hyperparameter_robustness.tex
\begin{table*}[t]
\centering
\caption{Average Success Rate On Meta-World with different RGB Loss Weights and Heatmap Standard Deviations.}
\label{tab:robustness_analysis}

\small
\setlength{\tabcolsep}{5pt}
\renewcommand{\arraystretch}{1.18}

\begin{tabularx}{1.0\textwidth}{
>{\raggedright\arraybackslash}p{0.13\textwidth}
*{6}{>{\centering\arraybackslash}X}
}
\toprule
\multirow{2}{*}{Hyperparameter}
& \multicolumn{3}{c}{RGB Loss Weight ($\lambda$)}
& \multicolumn{3}{c}{Heatmap Standard Deviation ($\sigma$)} \\
\cmidrule(lr){2-4}\cmidrule(lr){5-7}
& \makecell{0.1}
& \makecell{0.5}
& \makecell{0.9}
& \makecell{1.5}
& \makecell{2.5}
& \makecell{3.5} \\
\midrule
Avg. Suc. (\%)
& 92 & 89.1 & 89.1
& 89.1 & 89.7 & 86.9 \\
\bottomrule
\end{tabularx}

\vspace{-0.6em}
\end{table*}